\documentclass[journal,twoside,web]{ieeecolor}
\usepackage{tmi}
\usepackage{cite}
\usepackage{amsmath,amssymb,amsfonts}
\usepackage{algorithmic}
\usepackage{graphicx}
\usepackage{textcomp}
\usepackage{color}
\usepackage{colortbl}
\usepackage{algorithm}
\usepackage{subcaption}
\usepackage{float}
\usepackage{booktabs}
\usepackage{xspace}
\usepackage{epstopdf}
\newcommand{\etal}{et~al.\xspace}
\newcommand{\eg}{e.g.\xspace}
\newcommand{\ie}{i.e.\xspace}
\newcommand{\modelname}{DeReF\xspace}

\newcommand{\why}[1]{{\color{black}#1}}

\newcommand{\rev}[1]{{\color{black}#1}}
\newcommand{\revm}[1]{{\color{black}#1}}

\def\BibTeX{{\rm B\kern-.05em{\sc i\kern-.025em b}\kern-.08em
    T\kern-.1667em\lower.7ex\hbox{E}\kern-.125emX}}
\begin{document}
\title{Decouple, Reorganize, and Fuse: A Multimodal Framework for Cancer Survival Prediction}

\author{Huayi Wang, Haochao Ying, \IEEEmembership{Member, IEEE}, Yuyang Xu, Qibo Qiu, Cheng Zhang, Danny Z. Chen, \IEEEmembership{Fellow, IEEE}, Ying Sun, and Jian Wu, \IEEEmembership{Member, IEEE}
\thanks{This research was partially supported by National Natural Science Foundation of China under Grant No. 92259202 and No. 62476246, "Pioneer" and "Leading Goose" R\&D Program of Zhejiang under Grant No. 2025C02132, and GuangZhou City’s Key R\&D Program of China under Grant No. 2024B01J1301. (Corresponding Authors: Haochao Ying and Jian Wu)
}
\thanks{Huayi Wang and Yuyang Xu are with the College of Computer Science and Technology, Zhejiang University, Hangzhou 310012, China. They are also with the State Key Laboratory of Transvascular Implantation Devices of the Second Affiliated Hospital, Zhejiang University School of Medicine, and Transvascular Implantation Devices Research Institute, Hangzhou 310009, China, and Zhejiang Key Laboratory of Medical Imaging Artificial Intelligence, Hangzhou 310058, China. (e-mail: huayiwang@zju.edu.cn; xuyuyang@zju.edu.cn)}
\thanks{Haochao Ying and Jian Wu are with the State Key Laboratory of Transvascular Implantation Devices of the Second Affiliated Hospital, Zhejiang University School of Medicine, and Transvascular Implantation Devices Research Institute, Hangzhou 310009, China. They are also with the School of Public Health, Zhejiang University, Hangzhou 310058, China. Jian Wu is also with Zhejiang Key Laboratory of Medical Imaging Artificial Intelligence, Hangzhou 310058, China. (e-mail: haochaoying@zju.edu.cn; wujian2000@zju.edu.cn)}
\thanks{Qibo Qiu is with the China Mobile (Zhejiang) Research \& Innovation Institute, and the College of Computer Science and Technology, Zhejiang University, Hangzhou 310027, China. (e-mail: qiuqibo\_zju@zju.edu.cn)}
\thanks{Cheng Zhang and Ying Sun are with State Key Laboratory of Oncology in South China, Guangdong Key Laboratory of Nasopharyngeal Carcinoma Diagnosis and Therapy, Guangdong Provincial Clinical Research Center for Cancer, Sun Yat-sen University Cancer Center, Guangzhou 510060, China. (e-mail: zhangcheng@sysucc.org.cn; sunying@sysucc.org.cn)}
\thanks{Danny Z. Chen is with the Department of Computer Science and
Engineering, University of Notre Dame, Notre Dame, IN 46556 USA.
(e-mail: dchen@nd.edu)}
}

\maketitle

\begin{abstract}

Cancer survival analysis commonly integrates information across diverse medical modalities to make survival-time predictions. 
Existing methods primarily focus on extracting different decoupled features of modalities and performing fusion operations such as concatenation, attention, and \revm{Mixture-of-Experts (MoE)-based fusion.} 
However, these methods still face two key challenges: i) \rev{fixed} fusion schemes (concatenation and attention) can lead to model over-reliance on predefined feature combinations, limiting the dynamic fusion of decoupled features; ii) in MoE-based fusion methods, each expert network handles separate decoupled features, which limits information interaction among the decoupled features. 
To address these challenges, we propose a novel \underline{De}coupling-\underline{Re}organization-\underline{F}usion framework (DeReF), which devises a random feature reorganization strategy between modalities decoupling and dynamic MoE fusion modules.
Its advantages are: 
i) it increases the diversity of feature combinations and granularity, enhancing the generalization ability of the subsequent expert networks; 
ii) it overcomes the problem of information closure and helps expert networks better capture information among decoupled features. 
Additionally, we incorporate a regional cross-attention network within the modality decoupling module to improve the representation quality of decoupled features. 
Extensive experimental results on our in-house Liver Cancer (LC) and three widely used public datasets from \revm{The Cancer Genome Atlas (TCGA)} confirm the effectiveness of our proposed method.
\rev{Codes are available at https://github.com/ZJUMAI/DeReF.}
\end{abstract}


\section{Introduction}
\label{sec:intro}

Cancer survival analysis, a fundamental clinical task, estimates the probability of a specific event occurring over time, including mortality, disease progression, and recurrence~\cite{d2021methods,Greenhouse_Stangl_Bromberg_1989}.
This task not only helps physicians optimize treatment regimens and provides decision support for clinical care, but also helps patients better prepare for disease progression~\cite{Giordano_2021,sox2024medical}. \revm{In this context, cancer survival prediction aims to model patient-specific time-to-event outcomes by estimating survival probability or risk over a follow-up period, while explicitly accounting for censored cases where the event has not occurred by the last observation. Therefore, the goal is not only to assess whether an event may happen, but also to characterize when it is likely to occur, which is essential for risk stratification and personalized treatment planning~\cite{LERADEMACHER20211067,clark_survival_2003,P202262}.}
Medical data used for survival prediction usually presents diversity and heterogeneity, such as image data (\revm{\eg Magnetic Resonance Imaging (MRI)}), pathological data (\revm{\eg Whole Slide Imaging (WSI)}), genomic data (\revm{\eg RNA sequencing (RNA-seq)}), etc.
These modalities provide rich and complementary information.
\rev{For example, WSI reflects spatial histomorphological patterns like the structural features of tumors, while genomic profiling reveals molecular alterations without spatial context. Their integration offers a comprehensive view of tumor biology, which has shown potential to improve prognostic accuracy and support personalized treatment decisions~\cite{chen2022pan,Sabah_2021,Boehm_2022,vale2021long}.}
\rev{However, integrating such heterogeneous data remains a significant challenge, as tumors exhibit complexity across modalities with distinct formats, scales, resolutions, and semantic levels—from genetic variations to histopathological and radiological phenotypes~\cite{luo2025large,wang2025multimodal}. Directly merging these heterogeneous data will introduce semantic conflicts during subsequent cross-modal interactions, which may blur meaningful prognostic signals and degrade predictive performance~\cite{zhou2024multimodal,li2025application,duan2024deep}.}
A mainstream approach for multimodal-type prediction first decouples multimodal data and then fuses the outcomes from decoupled data for final prediction. 
\rev{It is worth noting that decoupling refers to the disentanglement of latent feature representations across modalities, aiming to extract separated features such as modality-specific and modality-shared components, thereby achieving a cleaner and more effective fusion process.}
Decoupling methods generally can be divided into two types.
The first type focuses on extracting and utilizing only modality-shared or modality-specific features~\cite{he2021multi,braman2021deep,zhou2019deep,9366692}. Such a method may cause the model to overlook potential interaction among decoupled features, limiting the overall performance. 
The second type seeks to extract and use all decoupled features through modality disentanglement~\cite{liu2022disentangled,zhang2022learning,ouyang2021representation}. But, such a method still faces challenges in effectively extracting decoupled features. For instance, modality-shared features are often obtained by concatenating unimodal features and processing them in fully connected layers or by applying cross-attention mechanisms~\cite{CFDL,zhou2024cohort}. This type of methods favors the consideration of inter-modal interaction and weakly learns intra-modal interaction. Neglecting such interactions may lead to inadequate representations of decoupled features.
In terms of fusing decoupled features, the current fusion methods perform concatenation, attention, or MoE. For example, Wei \etal ~\cite{wei2025robust} directly concatenated modality-shared and modality-specific features, and fed them into a fully connected layer for prediction. Zhang \etal ~\cite{PIBD} put decoupled features into a Transformer~\cite{vaswani2017attention} to make predictions. 
\rev{However, these methods employ fixed decoupled feature fusion strategies---directly concatenating decoupled features at predefined positions---which may lead to a rigid modeling of feature relationships. This rigidity restricts the model’s ability to dynamically capture potential interactions among different decoupled sub-features.}
Some fusion methods used the MoE model~\cite{jacobs1991adaptive,2021GShard,fedus2022switch}: Each expert network mines information of different decoupled features independently, after which a gating unit is applied to conduct dynamic fusion of different expert branches~\cite{CFDL}. 
\rev{Compared to the direct concatenation method, the MoE method dynamically assigns importance weights to different feature components, enabling more balanced and effective integration~\cite{wang2025decoupled}.}
However, these methods suffer from the information closure problem, where each expert network considers only specific information on decoupled features and lacks consideration of useful information among decoupled features.

\begin{figure}[t]
  \centering
   \includegraphics[width=\linewidth]{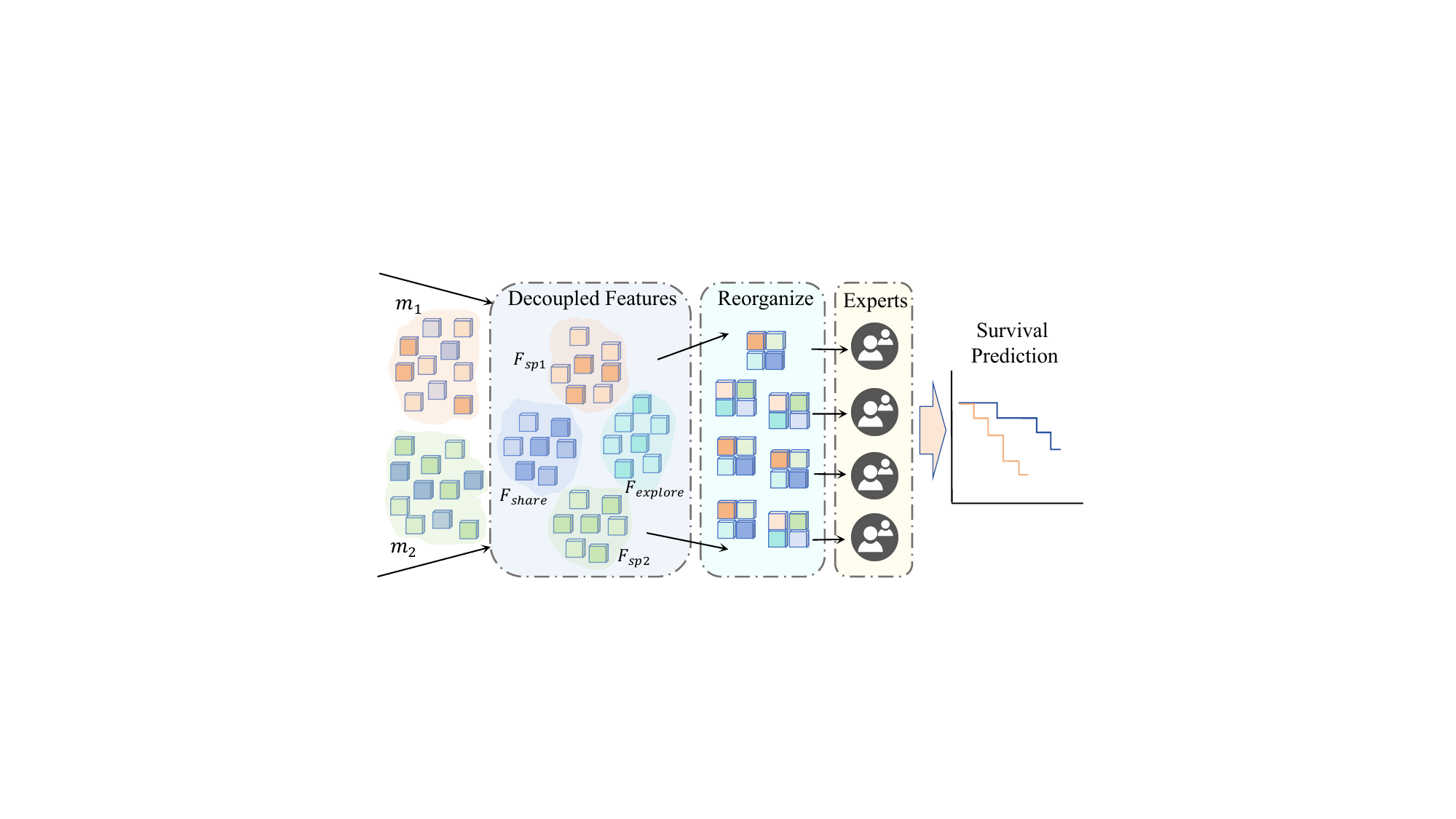}
   \caption{Model Architecture Diagram. The random reorganization of decoupled features is conducive to 1) increasing the diversity of feature combinations and granularity, enhancing the generalization performance of expert networks, and 2) increasing the interaction of different decoupled features in expert networks.  
   }
   \label{fig:intro}
    \vspace{-2ex}
\end{figure}



\rev{In this paper, we introduce a novel \underline{De}coupling-\underline{Re}organization-\underline{F}usion Framework (\modelname). It is designed to overcome key limitations in current methods, namely the mediocre quality of decoupled features and the rigid learning of their fusion relations. The overall framework is shown in Fig.~\ref{fig:intro}.}
First, we propose a regional cross-attention network in the feature decoupling module. 
\rev{By leveraging distinct sub-regions of the global attention matrix (an intermediate cross-attention product), this network enhances the modeling of intra- and inter-modality relationships, thereby improving decoupled feature quality.}
\rev{Furthermore, at the stage of fusing decoupled features, we design a random feature reorganization algorithm coupled with an MoE fusion module comprising multiple expert networks. Specifically, the random feature reorganization algorithm diversifies the granularity and the combination of decoupled features, enhancing the generalization capability of expert networks.} It can also alleviate information closure, enabling expert networks to better capture interactions among decoupled features.
\rev{Finally, within the MoE model, a gating network is employed to exploit the global information of fused decoupled features. It generates dynamic weight scores, which are then used to compute a weighted sum of the expert networks' outputs.}
Our main contributions are summarized as follows:

\begin{itemize}

    \item \rev{
    We propose DeReF, which develops a novel decouple-reorganize-fusion paradigm to achieve effective integration of heterogeneous medical multimodal data for cancer survival prediction.
    }
    \item \rev{We devise two core components: a regional cross-attention algorithm for high-quality feature decoupling, and a random feature reorganization strategy for enhanced generalization and interaction modeling.}

    \item 
    \revm{Our method achieves the highest average C-Index scores of 0.674 and 0.683 on LC and TCGA datasets, respectively. In particular, the best-performing setting on LC dataset improves upon the strongest baseline by 2.1\%, demonstrating its superiority. 
    }

\end{itemize}

\begin{figure*}
    \centering
    \includegraphics[width=1\linewidth]{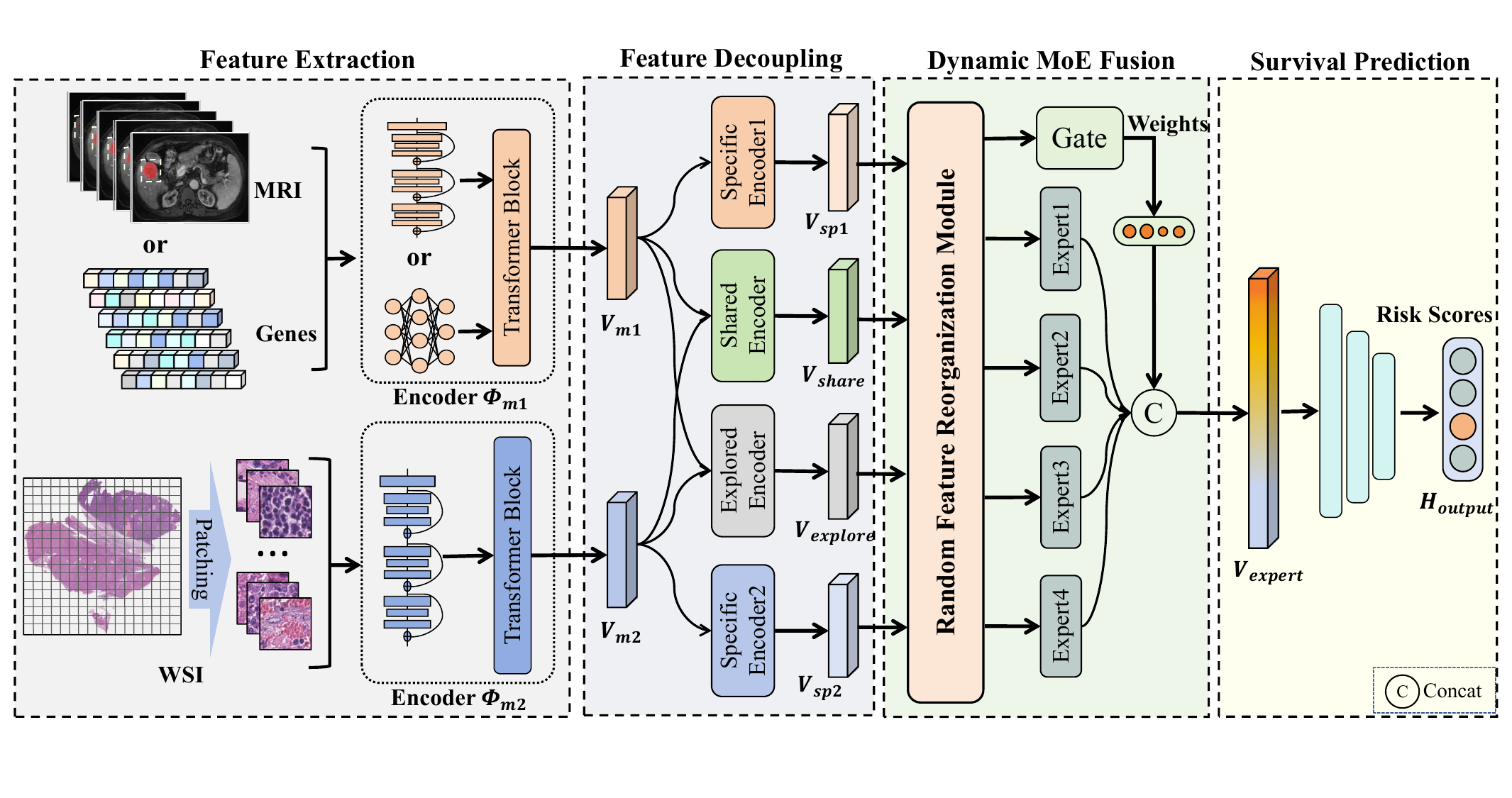}
    \caption{An overview of our proposed framework. It consists of four modules: the Feature Extraction Module, Feature Decoupling Module, Dynamic MoE Fusion Module (containing Random Feature Reorganization and an MoE model), and Survival Prediction Module.}
    \label{fig:framework}
    \vspace{-2ex}
\end{figure*}

\section{Related Work}\label{sec:relat}

\subsection{Multimodal Feature Decoupling}
In multimodal learning, the main goal of feature decoupling is to separate features that are similar and complementary in multimodal data. 
This type of approach assumes that features from different modalities (e.g., MRI and WSI) are both heterogeneous and intrinsically related to each other. 
Previous work explored three distinct strategies: utilizing only modality-shared~\cite{9366692,braman2021deep} or modality-specific features~\cite{he2021multi,zhou2019deep}, and utilizing all decoupled features~\cite{liu2022disentangled,yang2022disentangled,zhang2022learning,ouyang2021representation,PIBD,2022_Cheng}.
For example, Ning \etal ~\cite{9366692} established a bidirectional mapping between the original space and the multimodal shared space, effectively obtaining a modality-shared representation. He \etal~\cite{he2021multi} proposed the Modality-Specific Attention Network (MSAN), which incorporates a multiscale and region-guided attention module to extract and exploit modality-specific features.
Zhou \etal~\cite{zhou2019deep} introduced the Deep Latent Multi-modality Dementia Diagnosis (DLMD) framework, based on deep non-negative matrix factorization (NMF), to learn potential representations from multimodal data for dementia diagnosis.
However, shared and specific features individually often tend to overlook intra-modal relationships, failing to learn comprehensive representations of multimodal data.
Cheng \etal~\cite{2022_Cheng} developed a Multi-Modal Disentangled Variational Auto-Encoder that disentangles latent representations into shared and specific components, ensuring validity through cross-modal reconstruction and shared-specific loss. 
However, it directly multiplies the variances of different feature distributions to obtain modality-shared features. This method may lead to information loss, especially in regions with low variance and lacking intermodal interaction.

To address the above limitations, we apply \rev{Multi-Layer Perceptron (MLP)} to extract modality-specific features, and introduce a novel regional cross-attention network to extract modality-shared and modality-explored features (see Section~\ref{sec:fd}). Notably, during the feature decoupling process, the cross-attention computation enables the model to comprehensively mine intra- and inter-modality information, thus improving the representation quality of decoupled features.


\subsection{Multimodal Feature Fusion}

The current multimodal feature fusion methods are mainly divided into direct connection (concatenation), attention-based, or MoE methods.
\rev{For instance, works~\cite{CMTA, MCAT, SurvPath} employs attention-based fusion methods to enhance interaction between modality features. Xiong et al.~\cite{xiong2024mome} proposed a progressive fusion approach based on MoE, where expert modules perform feature interactions through operations such as concatenation, averaging, and attention mechanisms. However, these approaches fail to decouple and separate modality feature representations, which may lead to issues such as semantic conflicts during feature interactions.}
Several studies~\cite{2022_Cheng,wei2025robust,PIBD,zhou2024cohort} concatenated modality-shared and modality-specific features, and fed them to fully connected or attention layers to produce predictions. But, we argue that these fusion methods can lead to over-reliance of the model on fixed feature combinations and lack ability to dynamically fuse decoupled features.
Liu \etal~\cite{CFDL} introduced a dynamic fusion method based on MoE, where modality-shared and modality-specific features are sent to four expert networks individually. In addition, a gating network is used to assign weight scores and fuse the features by weighted concatenation. 
However, each expert dealing with a separate decoupled feature may lack the extraction of relational information among the decoupled features, affecting the learning quality of the representations within the expert networks.

We propose a random feature reorganization strategy before applying the MoE model to address the above issues, by randomly segmenting and recombining the decoupled features. This approach not only reduces the over-reliance of the expert networks on fixed feature combinations and enhances the generalization ability of the expert networks, but also allows the expert networks to fully exploit the synergistic relationships among decoupled features.

\section{Method}

\subsection{Overall Framework}
Our framework is shown in Fig.~\ref{fig:framework}. The framework contains three main modules: a feature extraction module, a feature decoupling module, and a decoupled feature fusion module.
First, we use $X_{\text{m1}}$ and $X_{\text{m2}}$ to represent the data of two modalities. 
The pre-processed multimodal data are fed into the corresponding encoders (\revm{$\Phi_{\text{m1}}$, $\Phi_{\text{m2}}$}) to obtain features ($V_{\text{m1}}$, $V_{\text{m2}}$), respectively. 

\rev{The encoders consist of feature extraction network and lightweight Transformer Block same as TransMIL~\cite{TransMIL}.}
Second, to reduce information interference between heterogeneous modalities, the extracted features are fed into the feature decoupling module to obtain four decoupled features, including modality-shared, modality-specific, and modality-explored features. Modality-specific and modality-explored decoupled features are obtained by our proposed novel cross-attention method, improving decoupled representation quality. Third, the decoupled features are fused using a dynamic fusion method based on MoE. A random feature reorganization strategy is developed to improve the generalization ability of MoE. Finally, a fully connected layer is used to output predictions.
We develop two loss functions $\mathcal{L}_{\text {surv }}$ and $\mathcal{L}_{\text {dis }}$ to supervise the final prediction and regulate the distance between the decoupled features.

\subsection{Feature Extraction}
\label{sec:fe}
Since our framework is experimented on different datasets, we let $X_{\text{m1}}$ represent MRI or Genomic profiles, and $X_{\text{m2}}$ represent WSI. 
For the MRI data, we first crop and resize the \revm{Region of Interest (ROI)} of tumor according to the \why{manual} segment labels, then extract the tumor features \rev{by 3D ResNet50 network}, and finally use global average pooling to obtain MRI features. 

For the genomic profiles, following previous works~\cite{CMTA,MCAT,MOTCat}, we extract six genomic sub-sequence features using \revm{Self-Normalizing Neural Network (SNN)}~\cite{SNN}, a widely used network for modeling genomic features. $\widehat{V}_{\text{m1}}$ represents the features of MRI ($\in \mathbb{R}^{1\times C_1}$) or Genomic profiles ($\in \mathbb{R}^{6\times C_1}$). 

\revm{For the WSI data, due to its extremely high resolution, we perform offline (pre-computed) WSI cropping and feature extraction by encoding the resulting patches once with a frozen pretrained backbone and caching the embeddings, which are then directly loaded during training instead of repeatedly processing the raw WSIs.}
Following previous works~\cite{CMTA,MCAT,MOTCat}, \revm{we adopt the preprocessing pipeline from Clustering-constrained Attention Multiple instance learning (CLAM)~\cite{CLAM}} to crop each WSI into a series of non-overlapping $256\times256$ patches at $20\times$ magnification level.
Then, ResNet50 (\rev{Load public weight pretrained on ImageNet dataset}) is used to
extract features $\widehat{V}_{\text{m2}} \in \mathbb{R}^{I\times C_1}$.

Finally, considering the dimensional alignment of different modality features, we use Nystrom Attention~\cite{xiong2021nystromformer} to obtain the class tokens as the final feature representations $V_{\text{m1}} \in \mathbb{R}^{1\times C_1}$ and $V_{\text{m2}}\in \mathbb{R}^{1\times C_1}$. The class token serves as a global representation derived from the aggregated features of  $\widehat{V}_{\text{m1}}$ or $\widehat{V}_{\text{m2}}$. Here, the patch number (denoted as $I$) corresponds to the total number of non-overlapping patches extracted from each WSI, and $C_1$ represents the embedding vector length.

It is worth noting that in the encoder, except for the feature extraction of WSI data is off-line, the rest of the network is trained from scratch along with the whole framework.
\begin{algorithm}[t] 
\caption{Regional Cross-Attention Algorithm.} 
\label{alg:Framwork} 
\begin{algorithmic}[1] 
\REQUIRE ~~\\ 
Modality features: $\{V_{\text{m1}}$, $V_{\text{m2}}\} \in \mathbb{R}^{1 \times C_1}$.\\
\ENSURE ~~\\ 
Decoupled features: $\{V_{\text{share}}$, $V_{\text{explore}}\} \in \mathbb{R}^{1\times C_2} $.
\STATE \textbf{Embedding Projection:}\\
$v_{\text{m1}} = \text{FC}_1(V_{\text{m1}})\in \mathbb{R}^{1 \times C_2}$,
$v_{\text{m2}} = \text{FC}_2(V_{\text{m2}})\in \mathbb{R}^{1 \times C_2}$; \\
// $\text{FC}_1, \text{FC}_2$ are learnable linear projections

\label{code:fram:extract }

\STATE \textbf{Feature Concatenation:}\\
$X_1 = [v_{\text{m1}}, v_{\text{m2}}] \in \mathbb{R}^{1 \times C_1}$,
$X_2 = [v_{\text{m2}}, v_{\text{m1}}] \in \mathbb{R}^{1 \times C_1}$;

\label{code:fram:trainbase}

\STATE \textbf{Attention Matrix Computation:}\\
$\mathbf{M} = X_1^\top \cdot X_2 \in \mathbb{R}^{C_1 \times C_1}$;
\label{code:crossattention}

\STATE \textbf{Attention Matrix Partition:}\\
// Obatin four submatrices\\
$\mathbf{M}=\begin{bmatrix} \mathbf{M}_{\text{m1m2}} & \mathbf{M}_{\text{m1m1}}\\ \mathbf{M}_{\text{m2m2}} & \mathbf{M}_{\text{m2m1}}\\
\end{bmatrix}$, \\ where $\{\mathbf{M}_{\text{m1m2}},\mathbf{M}_{\text{m1m1}},\mathbf{M}_{\text{m2m2}},\mathbf{M}_{\text{m2m1}}\} \in \mathbb{R}^{C_2 \times C_2};$

\label{code:submatrix}
\STATE \textbf{Regional Cross-Attention Computation:}\\
$ \begin{bmatrix} v_{\text{m1}}^{\text{(m)}} , v_{\text{m2}}^{\text{(s)}} \end{bmatrix} = \begin{bmatrix} v_{\text{m1}} ,v_{\text{m2}} \end{bmatrix} \cdot \text{Softmax} ( \begin{bmatrix}
 \mathbf{M}_{\text{m1m2}}, \mathbf{M}_{\text{m2m2}} \end{bmatrix}) \in \mathbb{R}^{1 \times C_2}$ ,\\
$ \begin{bmatrix} v_{\text{m2}}^{\text{(m)}} , v_{\text{m1}}^{\text{(s)}} \end{bmatrix} = \begin{bmatrix} v_{\text{m2}} ,v_{\text{m1}} \end{bmatrix} \cdot \text{Softmax}(\begin{bmatrix} \mathbf{M}_{\text{m1m2}}, \mathbf{M}_{\text{m1m1}} \end{bmatrix}^\top) \in \mathbb{R}^{1 \times C_2}$; \label{code:attnres}

\STATE \textbf{Attention Scores Fusion:} \\
// Averaging attention scores to output\\

$V_{\text{output}} = \frac{1}{2} \left(
\begin{bmatrix}
v_{\text{m1}}^{(\text{m})}, v_{\text{m2}}^{(\text{s})}
\end{bmatrix} +
\begin{bmatrix}
v_{\text{m2}}^{(\text{m})}, v_{\text{m1}}^{(\text{s})}
\end{bmatrix}
\right) \in \mathbb{R}^{1 \times C_2}$;
 \label{code:finalres}

\RETURN $V_{\text{output}}$. 
\end{algorithmic}
\end{algorithm}

\subsection{Feature Decoupling}
\label{sec:fd}
In multimodal learning, due to the heterogeneous gap between different modalities, decoupling can help the model to better find the consistency and complementarity between modalities in the feature space, reduce the information interference generated by modal fusion~\cite{Wang2024BDPartNetFD}. 
Therefore, we decouple two modality features into four parts, including two modality-specific, one modality-shared, and one modality-explored features. The modality-specific feature is used to preserve the specific information of the two modality data. 
\why{The modality-shared feature contains explicit similarities between modalities, while modality-explored feature contains implicit supplementary information derived from inter-modality interactions.
Essentially, the core motivation for introducing modality-explored feature is to facilitate the model in capturing more inter-modality interactions through distance control, thereby improving its predictive performance.}
We illustrate modality-shared and modality-explored features with two examples:

Example 1: The low expression of E-cadherin exhibits an explicit association with poor tumor cell differentiation (E-cadherin is a key protein for cell-cell adhesion, and its low expression leads to disorganized cell arrangement~\cite{van2008cell}). Thus, the relationships between low expression of E-cadherin and disorganized cell arrangement can be viewed as modality-shared feature.

Example 2: High expression of TGF-$\beta$ in tumor microenvironment may trigger complex biological effects, such as immune suppression, fibrosis, and angiogenesis. However, TGF-$\beta$ high expression does not directly cause fibrosis or immune cell infiltration in WSIs but indirectly influences the tumor microenvironment through a series of complex biological processes~\cite{zhang2024tertiary} (\eg, signal transduction, cell-cell interactions). So the relationships between TGF-$\beta$ and tumor microenvironment are nonlinear and implicit, which can be viewed as modality-explored feature.


\revm{In particular}, the specific features $V_{\text{sp1}}, V_{\text{sp2}}$ are extracted using two MLP layers.
Modality-shared and modality-explore features $V_{\text{share}},V_{\text{explore}}$ can be obtained by proposed regional cross-attention algorithm, see Algorithm~\ref{alg:Framwork}. In this algorithm, the $V_{\text{output}}$ represents $V_{\text{share}}$ or $V_{\text{explore}}$, and $\{V_{\text{sp1}},V_{\text{sp2}},V_{\text{share}},V_{\text{explore}}\}\in \mathbb{R}^{1\times C_2}$.
\why{The attention matrix $M$ is obtained by multiplying the feature embeddings of two modalities with different concatenation orders. This design enables direct extraction of both inter- and intra-modality region matrices from M (\eg, $\begin{bmatrix} \mathbf{M}_{\text{m1m2}}, \mathbf{M}_{\text{m2m2}} \end{bmatrix})$. 
Compared with work~\cite{zhou2024cohort}, which only considers cross-modal attention region ($\mathbf{M}_{m1m2}$), our algorithm can analyze both inter- and intra-relationships of features, allowing the integration of core information from both modalities. }
Subsequently, we measure the similarity among $\{V_{\text{sp1}},V_{\text{sp2}},V_{\text{share}},V_{\text{explore}}\}$ using the distance function $\text{Dis}(V_i,V_j)$ to formulate the decoupling loss: 
\begin{equation}
   \begin{split}
   \mathcal{L}_{\text{dis}}&=\text{Dis}(V_{\text{sp1}},V_{\text{share}})+\text{Dis}(V_{\text{sp2}},V_{\text{share}})\\&+\text{Dis}(V_{\text{sp1}},V_{\text{explore}})+\text{Dis}(V_{\text{sp2}},V_{\text{explore}})\\
    &+\text{Dis}(V_{\text{share}},V_{\text{explore}})-\text{Dis}(V_{\text{sp1}},V_{\text{sp2}}).
    \end{split}
    \label{eq:loss_dis}
\end{equation}
The distance between $V_{\text{sp1}}$ and $V_{\text{sp2}}$ should be maximized to ensure their distinctiveness. Furthermore, since $V_{\text{share}}$ represents the shared features across different modalities, it is required to maintain close associations with \{$V_{\text{sp1}}$,$V_{\text{sp2}}$\}. 
\rev{Additionally, as a semantic extension and boundary exploration of $V_{\text{share}}$, $V_{\text{explore}}$ is designed to preserve correlation with the modality-shared features while capturing implicit associations not explicitly modeled in the primary decoupling path. To encourage such behavior, its distances to $\{V_{\text{sp1}},V_{\text{sp2}},V_{\text{share}}\}$ are minimized, which facilitates the learning of latent interactions and promotes coherence within the decoupled feature space.  }

\rev{In this study, we select several commonly used distance measurement formulas as distance functions, including L1 distance (L1), Kullback-Leibler divergence (KL), Cosine similarity (CoS), and Mean Square Error (MSE). Based on the results from the ablation experiments (Table~\ref{tab:distance}), this paper adopts MSE as core distance function.}

It is worth noting that although $V_{\text{share}}$ and $V_{\text{explore}}$ employ the same framework, they are initialized with different parameters. Moreover, their distances are constrained by the decoupling loss, which endows them with distinct semantic meanings.


\subsection{Feature Reorganization}
\label{sec:fr}
The idea of random feature reorganization is derived from ShuffleNet~\cite{zhang2018shufflenet}. In that work, group convolution allows each group's features to be learned independently. This leads to strong intra-group information interaction but weak inter-group information interaction. The channel shuffle enhances information flow across groups by disrupting the channel order and allowing features to be redistributed across groups.

\begin{figure}
  \centering
   \includegraphics[width=0.7\linewidth]{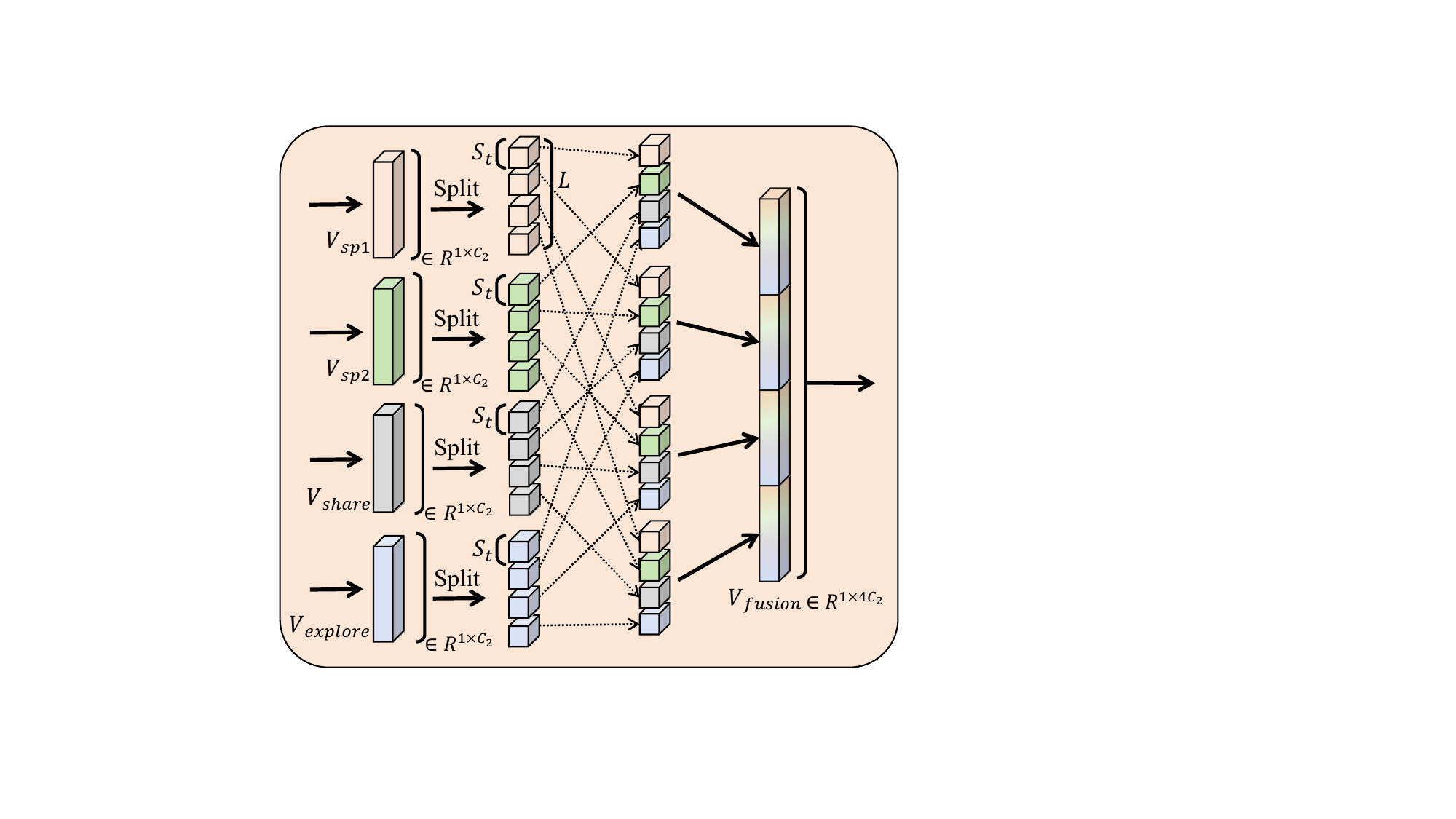}
   \caption{Illustration of Feature Reorganization Module. The reorganized decoupled features are fed into each of the four subsequent expert networks.}
   \label{fig:reorg}
    \vspace{-2ex}
\end{figure}

Slightly different from channel shuffle, we first split each decoupled feature into a number of equal sub-features, after which sub-features from different decoupled features are combined (see Fig.~\ref{fig:reorg}). 
Specifically, we first set $n$ segments $S=\{s_t\}_{t=1}^{n}$, \ie, divide each decoupled feature $\{V_o\}_{o=1}^4 \in \mathbb{R}^{1\times C_2}$ into $L$ equal feature segments $\{v_{o,l}\}_{l=1}^L$, where $L=C_2//s_t$. We use $o=\{1,2,3,4\}$ to represent feature names \{sp1,sp2,share,explore\} respectively. After that, each feature segment is reorganized, \ie, $V_{l}^{'}=\text{Concat}(v_{1,l},v_{2,l},v_{3,l},v_{4,l})$. Finally, the reorganized feature segments $\{V_{l}^{'}\}_{l=1}^L$ are concatenated along the channel dimension (obtain $V_{\text{fusion}}=\text{Concat}(\{V_{l}^{'}\}_{l=1}^L)$) and fed into the subsequent expert networks. 
Notably, the segment value $s_t$ is randomly chosen at every forward step.

\rev{We argue that random feature reorganization is applicable only to decoupled features, rather than the raw modality features ($V_{\text{m1}}$, $V_{\text{m2}}$). This is because the decoupling process separates the raw modality features into distinct components such as modality-shared and modality-specific representations. This separation establishes a clearer learning pathway for cross-modal interactions. The ablation study (Table~\ref{tab:ablation}, W/o Dec) further validates this claim.}

In conclusion, we use random feature reorganization for two purposes. First, it enhances the robustness of the expert networks. Compared to directly concatenating the four decoupled features, random feature reorganization breaks the fixed positional relationships among the original decoupled features. This allows the neurons in expert networks to avoid over-relying on specific positional features of certain decoupled features and gradually learn common information among the decoupled features. This approach effectively prevents expert networks from overfitting to certain features and reduces the model's dependence on specific feature combinations or orders, resulting in improved generalization ability.
Second, it improves the ability of expert networks to capture potential interaction information. Random feature reorganization rearranges the decoupled features at a finer granularity, allowing each expert network to focus more on local relationships and details among the decoupled features. This enables the model to learn potential interaction information from small-scale feature combinations.

\subsection{Dynamic MoE Fusion}
\label{sec:moe}
Currently, sparse MoE~\cite{hwang2023tutel,riquelme2021scaling} is commonly used, selecting the TopK expert networks for feature learning and dynamic feature fusion. 
However, random feature reorganization changes the original positional relationships of decoupled features, making it challenging for Sparse MoE to fully capture the latent information among shuffled features. To mitigate this issue, we propose a dynamic dense MoE~\cite{nie2021evomoe,pan2024dense} for dynamically fusing decoupled features, where all expert networks are activated. Each expert network can capture potential relationships among decoupled features from different aspects, similar to the multi-head self-attention mechanism. \rev{Furthermore, in contrast to traditional MoE fusion modules~\cite{CFDL}, where each expert network processes only a specific and fixed decoupled feature, our design enables each expert to process all decoupled features. This enhances the model's capabilities of information interaction and mitigates the issue of information closure.}
The gating network, meanwhile, considers the global information of the features and generates dynamic weights based on the input, thereby achieving a dynamic fusion of output features.
Specifically,  we have a set of $N$ expert networks $\{E_1,...,E_N\}$ and a gating network $G$. Features $V_{\text{fusion}} \in \mathbb{R}^{1\times 4C_2}$ will pass through the gate network and all expert models simultaneously. The gating network will output the weight scores of each expert:
\begin{equation}
    g_i(V_{\text{fusion}}) = \operatorname{Softmax}(W\cdot V_{\text{fusion}})_i,\quad i \in \{1, \dots, N\},
    \label{eq:gate}
\end{equation}
where $W$ denotes learnable weight matrix in the gating network and $i$ denotes expert index.
Each expert network $E_i$ consists of two MLP layers. The final output of the dense MoE $V_{\text{expert}} \in \mathbb{R}^{1\times4C_2}$ is obtained by weighted concatenation of the outputs of all the experts:
\begin{equation}
     V_{\text{expert}} = \text{Concat}(\{g_i(V_{\text{fusion}}) \cdot E_i(V_{\text{fusion}})\}_{i=1}^N).
     \label{eq:weightsum}
\end{equation}
Finally, we use one fully connected layer and sigmoid function to predict the final risk scores $H_{ouput}$: 
\begin{equation}
    H_{\text{output}} = \text{Sigmoid}(\text{MLP}(V_{\text{expert}}))).
    \label{eq:output}
\end{equation}


\subsection{Loss Function}\label{sec:lf}

Following previous work~\cite{zhou2024cohort}, we convert the original event time regression task into a time period-based classification task. Each sample can be defined as $\{H_{\text{output}},c,n\}$, where $H_{\text{output}} = \{h_1,\dots h_n\}$ represents the hazard output probability, $c$ represents the censored state, and $n$ represents the class label of time period. In addition, the discrete survival function $f_{\text {surv }}(H_{\text{output}},n)=
\prod_{j=1}^n\left(1-h_j\right)$. We use negative log-likelihood (NLL) with censorship to supervise the survival prediction:
\begin{equation}
    \begin{aligned}
\mathcal{L}_{\text {surv }}= & -c \log \left(f_{\text {surv }}(H_{\text{output}}, n)\right) -(1-c) \log \left(h_n\right)
 \\& -(1-c) \log \left(f_{\text {surv}}(H_{\text{output}}, n-1)\right).
\end{aligned}
\label{eq:loss_sur}
\end{equation}
Finally, the total loss function of our framework is:
\begin{equation}
    \mathcal{L} = \mathcal{L}_{\text {surv }}+ \alpha \mathcal{L}_{\text{dis}},
    \label{eq:totalloss}
\end{equation}
where $\alpha$ is the balance factor.

\section{Experiments}
\subsection{Experimental Settings}
\noindent \textbf{Datasets.} We conduct extensive experiments on our own LC Dataset and on three public TCGA datasets to verify the effectiveness of our proposed model. 
\rev{The LC Dataset contains 160 pairs of diagnostic MRI and WSI with ground-truth survival outcome. More statistical information of patient data in LC dataset is shown in Table~\ref{tab:lc_char}.}
\begin{table*}[h]
\centering
\caption{Patient characteristics of the in-house liver cancer (LC) dataset.}
\label{tab:lc_char}
\begin{tabular}{|l|p{9cm}|}
\hline
Characteristic & Value \\
\hline
Number of patients & 160 \\
\hline
Age, median (range) & 58 (26--78) \\
\hline
Sex, male, $n$ (\%) & 121 (75.6\%) \\
\hline
Microvascular invasion grade, $n$ (\%) & Grade 1: 129 (80.6\%); Grade 2: 21 (13.1\%); Grade 3: 10 (6.25\%) \\
\hline
Survival time, median (range), months & 19.2 (1.4--35.1) \\
\hline
MRI availability & 160 (100\%), including Arterial phase, Delay phase, Hepatobiliary phase, T1-weighted imaging, T2-weighted imaging (five sequences) \\
\hline
WSI availability & 160 (100\%), including \revm{Hematoxylin and Eosin (H\&E)}~\cite{lillie1954histopathologic} stained WSIs \\
\hline
Censored, $n$ (\%) & 2 (1.25\%) \\
\hline
Collection site(s) & Sun Yat-sen University Cancer Center, Guangzhou, China \\
\hline
\end{tabular}
\end{table*}
The TCGA\footnote{https://portal.gdc.cancer.gov} dataset contains paired WSIs and genomic profiles with ground-truth survival outcome, and we use the following cancer types: Bladder Urothelial Carcinoma (BLCA) (n=373), Uterine Corpus Endometrial Carcinoma (UCEC) (n=480), and Lung Adenocarcinoma (LUAD) (n=453).
 \rev{For genomic profiles in TCGA dataset, we utilize a preprocessed multi-omics data from existing mature research~\cite{chen2022pan}, which has been widely adopted in multimodal fusion survival analysis~\cite{CMTA,MCAT,MOTCat,zhou2024cohort}. It contains three types of genomic data: RNA sequencing (RNA-seq), copy number variation (CNV), and single nucleotide variation (SNV). The dataset is then categorized into six genomic subtypes: 1) Tumor Suppression, 2) Oncogenesis, 3) Protein Kinases, 4) Cell Differentiation, 5) Transcription, and 6) Cytokines and Growth.}


\noindent\textbf{Evaluation.} We perform 5-fold cross-validation 
and adopt the concordance index (C-Index)~\cite{harrell1982evaluating} to measure the performance of correctly ranking the predicted patient risk scores with respect to overall survival. Moreover, Kaplan-Meier \cite{kaplan1958nonparametric} analyses is utilized to assess the significance of differences in survival predictions between high- and low-risk groups.

\noindent\textbf{Implementation Details.} 
Our framework is implemented by PyTorch running on a single NVIDIA GTX 3090 GPU. During training, we adopt Adam optimizer with the initial learning rate of $5\times 10^{-4}$ and weight decay of $1\times 10^{-5}$. \why{Our framework is trained for 30 epochs}.
We set $C_1=256, C_2=128, S=\{2,8,16,32,64\}$, balance factor $\alpha=1$, expert number $N=4$. \rev{In addition, in the main and ablation experiments, we set the same random seed to ensure the reproducibility and fairness of the experiment.}

\subsection{Comparison with Known Methods}
\label{sec:mainresult}

We compare our proposed methods with both unimodal and multimodal state-of-the-art methods.

\noindent\textbf{Unimodal Baselines.} We evaluate our \modelname model against the classic model ResNet3d~\cite{Resnet} for MRI modality in the LC dataset, two state-of-the-art Multiple Instance Learning (MIL) methods AttnMIL~\cite{AttnMIL} and CLAM~\cite{CLAM} for WSI modality, and SNN~\cite{SNN} and SNNTrans~\cite{SNN} for genomic profiles modality.

\noindent\textbf{Multimodal Baselines.} On LC dataset, we compare our model with seven methods: MCAT~\cite{MCAT}, CMTA~\cite{CMTA}, MOTCat~\cite{MOTCat}, CFDL~\cite{CFDL}, CCL~\cite{zhou2024cohort}, MoME~\cite{xiong2024mome} \rev{and method based on Concat+MLP.} On three TCGA datasets, we add one additional method: PIBD~\cite{PIBD}, which is limited for WSIs and Genomic profiles. 

\noindent\textbf{Result Analysis on LC Dataset.}
From Table~\ref{tab:lc}, it is evident that MRI-based methods outperform WSI-based methods and even some multimodal methods. We argue that MRI provides functional information, such as tumor blood supply and metabolic activity, which are strongly associated with tumor aggressiveness and prognosis~\cite{pathak2004molecular}. In contrast, WSIs primarily capture histomorphological features, which, while indirectly reflecting tumor biology~\cite{amgad2024population}, are less direct than MRIs. Furthermore, some multimodal methods (MCAT, CMTA, MOTCat) may negatively impact performance due to inter-modality conflicts during fusion. Thus, decoupling-fusion methods (\eg, CCL, DeReF) often achieve superior results by reducing feature interference prior to fusion. Compared to CFDL and CCL methods, DeReF further enhances the quality of decoupled features and models diversified inter-modality relationships through regional cross-attention and random feature reorganization, achieving state-of-the-art performance.

\begin{table}[t]
\caption{C-Index (mean ± std) performance over our LC dataset. Imaging and Pathology refer to Imaging modality (MRIs) and pathological modality (WSIs). The best results and the second-best results are highlighted in \textbf{bold} and \underline{underlined}.}\label{tab:lc}
    \centering
    \begin{tabular}{l|cc|c}
        \toprule
        Model &  Imaging & Pathology & C-index \\
        \midrule
        ResNet3D \cite{Resnet}  & \checkmark   &  &
         0.645$\pm$0.036  \\
        \midrule
        AttnMIL \cite{AttnMIL} & &\checkmark &0.520$\pm$0.052 \\
        CLAM\_MB \cite{CLAM}   &    & \checkmark & 0.510$\pm$0.053  \\
        CLAM\_SB \cite{CLAM}   &    & \checkmark  & 0.518$\pm$0.041 \\
        TransMIL \cite{TransMIL}   &    & \checkmark  & 0.592$\pm$0.047 \\
        \midrule
        MCAT \cite{MCAT}   & \checkmark   & \checkmark  & 0.614$\pm$0.040 \\
        CMTA \cite{CMTA}   & \checkmark   & \checkmark  & 0.597$\pm$0.079 \\
        MOTCat \cite{MOTCat}   & \checkmark   & \checkmark  & 0.621$\pm$0.043 \\
        CFDL \cite{CFDL}   & \checkmark   & \checkmark  & 0.612$\pm$0.053 \\
         CCL \cite{zhou2024cohort} & \checkmark   & \checkmark  & 0.648$\pm$0.034
         \\
         MoME \cite{xiong2024mome}   & \checkmark   & \checkmark  & \underline{0.650$\pm$0.037} \\
        Concat+MLP  & \checkmark   & \checkmark  & 0.586$\pm$0.036 \\
         \modelname (ours)   & \checkmark   & \checkmark  & \textbf{0.671$\pm$0.029} \\
        \bottomrule
    \end{tabular}

\end{table}

\begin{table*}[t]
    \caption{C-Index (mean ± std) performance over three TCGA datasets. Genomic and Pathology refer to Genomic modality (Genomic profiles) and Pathological modality (WSIs), respectively. The best results and the second-best results are highlighted in \textbf{bold} and \underline{underlined}, respectively. 
    }
    \label{tab:tcga}
    \centering
    \begin{tabular}{l|cc|cccc}
        \toprule
        Model &  Genomic & Pathology & BLCA &UCEC &LUAD &Overall \\
        \midrule
        SNN \cite{SNN}   & \checkmark   &  & 0.618$\pm$0.022 &0.679$\pm$0.040 &0.611$\pm$0.047 & 0.636\\
        SNNTrans \cite{SNN}   & \checkmark  &  &0.659$\pm$0.032  &0.656$\pm$0.038 &0.638$\pm$0.022 & 0.651\\
        \midrule
        AttnMIL \cite{AttnMIL} &  &\checkmark &0.599$\pm$0.048 &0.658$\pm$0.036 &0.620$\pm$0.061 &0.626\\
        CLAM\_MB \cite{CLAM}   &    & \checkmark & 0.565$\pm$0.027 &0.609$\pm$0.082 &0.582$\pm$0.072 & 0.585\\
        CLAM\_SB \cite{CLAM}   &    & \checkmark  & 0.559$\pm$0.034 &0.644$\pm$0.061 &0.594$\pm$0.063 &0.599\\
        TransMIL \cite{TransMIL}   &    & \checkmark  & 0.575$\pm$0.034 &0.655$\pm$0.046 &0.642$\pm$0.046 &0.624\\
        \midrule
        MCAT \cite{MCAT}   & \checkmark   & \checkmark   &0.672$\pm$0.032 &0.649$\pm$0.043 &0.659$\pm$0.027 &0.660\\
        CMTA \cite{CMTA}   & \checkmark   & \checkmark   &0.660$\pm$0.026 &0.660$\pm$0.049 &\underline{0.669$\pm$0.033} &0.663\\
        MOTCat \cite{MOTCat}   & \checkmark   & \checkmark  & 0.673$\pm$0.037 &0.675$\pm$0.040 &0.661$\pm$0.039 &0.670\\
        CFDL \cite{CFDL}   & \checkmark   & \checkmark  & \underline{0.675$\pm$0.025} &0.679$\pm$0.020 & 0.664$\pm$0.030& \underline{0.673}\\
        PIBD \cite{PIBD}   & \checkmark   & \checkmark  & 0.595$\pm$0.061 &-- &-- &0.595\\
        CCL \cite{zhou2024cohort} & \checkmark   & \checkmark  &0.640$\pm$0.041 &\underline{0.686$\pm$0.042} &0.657$\pm$0.037 & 0.661\\
        MoME \cite{xiong2024mome} & \checkmark   & \checkmark  &0.674$\pm$0.020 &0.679$\pm$0.035 &0.660$\pm$0.038 &0.671 \\
        Concat+MLP & \checkmark   & \checkmark  &0.665$\pm$0.046 &0.652$\pm$0.047 &0.644$\pm$0.057 &0.654 \\
         \modelname (ours)   & \checkmark   & \checkmark  & \textbf{0.681$\pm$0.031} & \textbf{0.688$\pm$0.046}& \textbf{0.671$\pm$0.045}&\textbf{0.680}\\ 
        \bottomrule
    \end{tabular}

\end{table*}

\noindent\textbf{Result Analysis on TCGA Datasets.} \why{The experimental results on three TCGA datasets are presented in Table~\ref{tab:tcga}. As can be seen, compared with all unimodal methods, the proposed method achieves the best performance, demonstrating its effectiveness in fusing multi-modal data. Quantitatively, our method outperforms the best unimodal methods by 2.2\%, 0.9\%, and 3.3\% in terms of the C-Index metric across the three datasets, respectively. 
Notably, genomic data generally exhibit superior performance compared to histopathological data, probably because genomic features directly reflect tumor-associated molecular characteristics that are highly correlated with tumor aggressiveness and prognosis. For instance, specific genes such as TP53 and EGFR have been confirmed to be significantly associated with survival rates~\cite{donehower2019integrated,masuda2025impact}. While histopathological data contain rich morphological information (\eg, cellular atypia and tissue architecture), their high resolution poses challenges in extracting effective features due to the presence of both homogeneous and heterogeneous regions.
}

Furthermore, our method surpasses the best multimodal methods by 0.6\%, 0.2\%, and 0.2\% in C-Index across the three datasets, achieving an average performance of 0.680. Overall, the proposed method exhibits several advantages. For example, compared to the CCL method, which only considers inter-modal relationships to extract decoupled features, our regional cross-attention analyzes extra intra-modal relationships through decoupling high-quality features. Compared to CFDL method, where each expert analyzes separate decoupled features, our random feature reorganization algorithm enable the expert networks to extract more potential information among the decoupled features, thus improving the model's generalization ability. Additionally, compared to MoME method, which also employs MoE but it does not explicitly decouple features, our method decouples features into distinct components. This allows the model to better understand the commonalities and differences across modalities, thereby mitigating potential information redundancy during the fusion process.



\subsection{Ablation Studies}
We conduct ablation studies on four datasets to evaluate the rationality of our proposed model empirically. We first study the impact of the model components, following the effect of different distance measures in the decoupling loss and the random segment length in the feature reorganization module. 

\begin{table}[t]
    \caption{The ablation results after removing five key components: 1) the decoupled feature $V_{\text{explore}}$; 2) the decoupling module (Dec); 3) the Regional Cross-Attention (RCA) in decoupling module;  4) the Random Feature Reorganization (RFR) module;  5) the MoE module. }
    \centering
    \begin{tabular}{l|cccc}
        \toprule
        Variants &  LC & BLCA  &  UCEC & LUAD\\
        \midrule     
         W/o $V_{\text{explore}}$  & 0.646$_{\pm\text{0.054}}$    & 0.678$_{\pm\text{0.028}}$ & 0.685$_{\pm\text{0.058}}$   & 0.645$_{\pm\text{0.024}}$ \\  
         W/o Dec  & $0.637_{\pm\text{0.028}}$    & $0.664_{\pm\text{0.027}}$ & $0.630_{\pm\text{0.040}}$   & $0.645_{\pm\text{0.014}}$ \\  
          W/o RCA$^1$  & 0.645$_{\pm\text{0.044}}$    & 0.669$_{\pm\text{0.028}}$ & 0.677$_{\pm\text{0.032}}$    & 0.662$_{\pm\text{0.023}}$ \\
        W/o RCA$^2$  & 0.660$_{\pm\text{0.036}}$    & 0.673$_{\pm\text{0.017}}$ & 0.681$_{\pm\text{0.024}}$    & 0.667$_{\pm\text{0.034}}$ \\   
        W/o RFR  & 0.637$_{\pm\text{0.021}}$   & 0.660$_{\pm\text{0.037}}$  & 0.674$_{\pm\text{0.030}}$   & 0.664$_{\pm\text{0.020}}$ \\
        W/o MoE   & 0.617$_{\pm\text{0.025}}$   & 0.651$_{\pm\text{0.045}}$  & 0.679$_{\pm\text{0.038}}$   &0.658$_{\pm\text{0.024}}$ \\    
        \midrule
        \modelname  & \textbf{0.671}$_{\pm\textbf{0.029}}$   &  \textbf{0.681}$_{\pm\textbf{0.031}}$  & \textbf{0.688}$_{\pm\textbf{0.046}}$   &  \textbf{0.671}$_{\pm\textbf{0.045}}$ \\
        \bottomrule
    \end{tabular}

    \label{tab:ablation}
\end{table}

\begin{table}[t]
    \caption{The impact of different distance metrics, including L1 distance (L1), Kullback-Leibler divergence (KL), Cosine similarity (CoS) and Mean Square Error (MSE).}
    \centering
    \begin{tabular}{l|cccc}
        \toprule
        Metrics &  LC & BLCA  &  UCEC & LUAD\\
        \midrule
         L1 & 0.640$_{\pm\text{0.030}}$ & 0.650$_{\pm\text{0.023}}$ & 0.685$_{\pm\text{0.075}}$ & 0.670$_{\pm\text{0.039}}$ \\
        KL & 0.639$_{\pm\text{0.040}}$ & 0.664$_{\pm\text{0.001}}$ & 0.674$_{\pm\text{0.045}}$ & 0.669$_{\pm\text{0.040}}$ \\
        CoS & 0.610$_{\pm\text{0.068}}$ & 0.660$_{\pm\text{0.040}}$ & 0.668$_{\pm\text{0.056}}$ & 0.656$_{\pm\text{0.018}}$ \\
        MSE & \textbf{0.671}$_{\pm\textbf{0.029}}$ & \textbf{0.681}$_{\pm\textbf{0.031}}$ & \textbf{0.688}$_{\pm\textbf{0.046}}$ & \textbf{0.671}$_{\pm\textbf{0.045}}$ \\ 
        \bottomrule
    \end{tabular}

    \label{tab:distance}
\end{table}

\begin{table}[t]
        \caption{The impacts of different choice methods of Segment Values (SV), including choosing from fixed values and randomly choosing from $S=\{2,8,16,32,64\}$.}
    \centering
    \begin{tabular}{l|cccc}

        \toprule
        SV &  LC & BLCA &  UCEC  & LUAD \\
        \midrule
       2 & 0.633$_{\pm\text{0.023}}$ & 0.671$_{\pm\text{0.036}}$ & 0.651$_{\pm\text{0.061}}$ & \textbf{0.673}$_{\pm\textbf{0.039}}$ \\
        8 & 0.619$_{\pm\text{0.023}}$ & 0.663$_{\pm\text{0.029}}$ & 0.674$_{\pm\text{0.046}}$ & 0.669$_{\pm\text{0.040}}$ \\
        16 & 0.587$_{\pm\text{0.060}}$ & 0.670$_{\pm\text{0.028}}$ & 0.664$_{\pm\text{0.047}}$ & 0.668$_{\pm\text{0.030}}$ \\
        32 & 0.587$_{\pm\text{0.071}}$ & 0.675$_{\pm\text{0.025}}$ & 0.657$_{\pm\text{0.074}}$ & 0.650$_{\pm\text{0.054}}$ \\
        64 & 0.640$_{\pm\text{0.046}}$ & 0.676$_{\pm\text{0.029}}$ & 0.665$_{\pm\text{0.058}}$ & 0.665$_{\pm\text{0.039}}$ \\ 
        \midrule
        Random & \textbf{0.671}$_{\pm\textbf{0.029}}$ & \textbf{0.681}$_{\pm\textbf{0.031}}$ & \textbf{0.688}$_{\pm\textbf{0.046}}$ & 0.671$_{\pm\text{0.045}}$ \\ 
        \bottomrule
    \end{tabular}

    \label{tab:seglength}
\end{table}

\begin{table}[t]
    \caption{The impact of the number of experts on model performance, including the number 2, 4, 6, 8.}
    \centering
    \begin{tabular}{l|cccc}
       \toprule
        Number & LC & BLCA & UCEC & LUAD \\
        \midrule
        2 & 0.655$_{\pm\text{0.043}}$ & 0.659$_{\pm\text{0.025}}$ & 0.659$_{\pm\text{0.048}}$ & 0.666$_{\pm\text{0.023}}$ \\
        4 & \textbf{0.671}$_{\pm\textbf{0.029}}$ & \textbf{0.681}$_{\pm\textbf{0.031}}$ & \textbf{0.688}$_{\pm\textbf{0.046}}$ & 0.671$_{\pm\text{0.045}}$ \\
        6 & 0.631$_{\pm\text{0.021}}$ & 0.680$_{\pm\text{0.028}}$ & 0.672$_{\pm\text{0.033}}$ & \textbf{0.673}$_{\pm\textbf{0.023}}$ \\
        8 & 0.620$_{\pm\text{0.033}}$ & 0.667$_{\pm\text{0.029}}$ & 0.670$_{\pm\text{0.047}}$ & 0.663$_{\pm\text{0.028}}$ \\
        \bottomrule
    \end{tabular}

    \label{tab:num_exp}
\end{table}

\begin{table}[t]
    \caption{The impact of different decoupling modules on the normalized mutual information between decoupled features ($V_{\text{share}}$/$V_{\text{explore}}$) and labels, including Concatenation+MLP, Cross-Attention (CA), and Regional Cross-Attention (RCA).}
    \centering
    \begin{tabular}{l|cccc}
        \toprule
        Variants & LC & BLCA & UCEC & LUAD \\
        \midrule    
        Concat+MLP & 0.232/0.236 & 0.257/0.147 & 0.254/0.221 & 0.263/0.227 \\  
        CA      & 0.271/0.163 & 0.296/0.171 & 0.263/0.242 & 0.272/0.253 \\  
        RCA     & \textbf{0.352/0.294} & \textbf{0.329/0.289} & \textbf{0.325/0.264} & \textbf{0.296/0.254} \\
        \bottomrule
    \end{tabular}
    \label{tab:mi_results}
\end{table}

\begin{table}[h]
    \centering
    \caption{Three five-fold cross-validation results (mean $\pm$ std) on four datasets. The `Average' row shows the mean and standard deviation of the performance means from three runs.}
    \begin{tabular}{l|cccc}
        \toprule
        Rounds& LC & BLCA & UCEC & LUAD \\
        \midrule
        1  & 0.669$_{\pm\text{0.030}}$ & 0.678$_{\pm\text{0.022}}$ & \textbf{0.690$_{\pm\text{0.033}}$} & \textbf{0.675$_{\pm\text{0.031}}$} \\
        2  & 0.667$_{\pm\text{0.016}}$ & \textbf{0.683$_{\pm\text{0.021}}$} & 0.683$_{\pm\text{0.032}}$ & 0.669$_{\pm\text{0.040}}$ \\
        3  & \textbf{0.673$_{\pm\text{0.027}}$} & 0.676$_{\pm\text{0.034}}$ & 0.685$_{\pm\text{0.036}}$ & 0.672$_{\pm\text{0.037}}$ \\
        \midrule
        Average & 0.670$_{\pm\text{0.003}}$ & 0.679$_{\pm\text{0.004}}$ & 0.686$_{\pm\text{0.004}}$ & 0.672$_{\pm\text{0.003}}$ \\
        \bottomrule
    \end{tabular}
    \label{tab:3_5fold}
\end{table}

\begin{table}[t]
    \centering
    \caption{Repeated-test results (mean $\pm$ std) on four datasets. The `Average' row reports the mean and standard deviation of the performance means from three repeated tests.}
    \begin{tabular}{l|cccc}
        \toprule
        Rounds & LC & BLCA & UCEC & LUAD \\
        \midrule
        1 & 0.668$_{\pm\text{0.031}}$ & 0.679$_{\pm\text{0.036}}$ & 0.685$_{\pm\text{0.037}}$ & 0.670$_{\pm\text{0.029}}$ \\
        2 & \textbf{0.674$_{\pm\text{0.028}}$} & \textbf{0.683$_{\pm\text{0.018}}$} & \textbf{0.692$_{\pm\text{0.039}}$} & \textbf{0.673$_{\pm\text{0.030}}$} \\
        3 & 0.673$_{\pm\text{0.026}}$ & 0.680$_{\pm\text{0.030}}$ & 0.689$_{\pm\text{0.038}}$ & 0.669$_{\pm\text{0.047}}$ \\
        \midrule
        Average & 0.672$_{\pm\text{0.003}}$ & 0.681$_{\pm\text{0.002}}$ & 0.689$_{\pm\text{0.004}}$ & 0.671$_{\pm\text{0.002}}$ \\
        \bottomrule
    \end{tabular}
    \label{tab:multi_test}
\end{table}

\noindent\textbf{Impact of Model Components.} We individually remove five key modules from the proposed framework, including the decoupled feature $V_{\text{explore}}$, the decoupling module, the Regional Cross-attention algorithm in the decoupling module, the Random Feature Reorganization module, and the MoE module. As shown in Table~\ref{tab:ablation}, removing any of these modules will negatively affect the precision of the model. 

In particular, removing the decoupled feature $V_{\text{explore}}$ results in a noticeable performance drop on LC and LUAD datasets, with decreases of 2.5\% and 2.6\%, respectively. In this case, the additonal implicit information carried by $ V_{\text{explore}} $ proves to be particularly important for model performance. 
\rev{To verify that the proposed feature reorganization strategy is specifically effective for decoupled features, we remove the decoupling module. On all datasets, performance drops by more than 2\%, indicating that feature decoupling establishes a clearer learning pathway for cross-modal interactions constituted by feature reorganization.}
\rev{When Regional Cross-Attention is removed from the decoupling module, we use Concatenation+MLP (W/o RCA$^1$) or traditional CA~\cite{zhou2024cohort} (W/o RCA$^2$) as alternatives respectively. On four datasets, the model performance decreases by 2.6\%/1.1\%, 1.2\%/0.8\%, 1.1\%/0.7\%, and 0.9\%/0.4\%, showing that cross-attention effectively captures intra- and inter-modality interactions, achieving robust decoupled feature representations.}
Further, the performance of the model decreased by 3.4\%, 2.1\%, 1.4\%, and 0.7\% without the Random Feature Reorganization module. This proves that mining more interaction information among decoupled features is crucial to improving the performance of the model. 
Finally, the ablation results also validate the MoE's capability for dynamic feature fusion and global relationship modeling, enabling the model to capture more comprehensive information and thus improve overall performance.

\noindent\textbf{Impact of Distance Metrics.} We explore the effect of different distance metrics in the decoupling loss function. By default, we use Mean Squared Error to measure the distance between the decoupled feature representations. We also compare with L1 distance, Kullback-Leibler divergence, and Cosine similarity. As shown in Table~\ref{tab:distance}, using MSE metric empirically yields better results. \revm{Calculation details are provided in the Supplementary Materials.}

\noindent\textbf{Impact of Segment Length.} To verify the hypothesis that diverse feature combinations can enhance the performance of the model, we further conduct experiments with fixed segment lengths, as shown in Table~\ref{tab:seglength}. \why{It can be observed that on LUAD dataset, $S=2$ achieves a superior result of 0.673, indicating that finer-grained feature reorganization helps the model extract more meaningful features on this dataset. However, in most cases, the results of using fixed segment lengths show varying degrees of degradation. 
} Overall, these results confirm that random feature segmentation is able to enhance the diversity of feature combinations and granularity in most scenarios, thereby improving the model’s ability to explore potential feature correlations.

\noindent\textbf{Impact of Expert Numbers.} \rev{We explore the impact of varying the number of experts in the MoE module on model performance. Results are presented in Table~\ref{tab:num_exp}. We observe that the optimal number of experts lies at either 4 or 6. Using only 2 experts or as many as 8 leads to degraded performance. We hypothesize that too few experts result in insufficient modeling capacity to capture the diverse patterns across decoupled features, while too many experts introduce functional overlap among expert networks, thereby impairing model performance.}

\noindent\textbf{Quality Evaluation of Decoupled Features.} \rev{To validate the Regional Cross-Attention (RCA) can enhance the quality of decoupled features ($V_{\text{share}}$/$V_{\text{explore}}$), we conduct a quantitative experiment based on mutual information~\cite{kinney2014equitability} (\revm{calculation details are provided in the Supplementary Materials}). The core hypothesis is that a higher mutual information value between the decoupled features and the risk labels indicates that the features contain more discriminative information regarding patient risk status, thus reflecting higher feature quality. We compare two alternative interaction methods: Concatenation + MLP and traditional Cross-Attention (CA)~\cite{zhou2024cohort}, as shown in Table~\ref{tab:mi_results}. The results demonstrate that our RCA method, owing to its superior capability in capturing both intra- and inter-modality interactions, outperforms the baseline methods on all datasets.}

\noindent\textbf{Robustness Evaluation of DeReF.} \rev{To evaluate the robustness of our model, we conduct two systematic experiments. First, we perform three independent rounds of five-fold cross-validation, in which the samples are randomly split into train and test datasets in each round. 
As shown in Table~\ref{tab:3_5fold}, the C-index remains highly consistent across the three runs, with a standard deviation below 0.005, indicating that the model’s performance is insensitive to random data partitions.
Second, to examine the stability of the Random Feature Reorganization (RFR) algorithm during test, we apply the same trained model to the identical test dataset for three repeated trials. The results, presented in Table~\ref{tab:multi_test}, exhibit nearly identical predictions across different runs. The results show that RFR algorithm not only enables the model to be invariant to different feature arrangements, but also effectively promotes the model's ability to capture diversified feature relationships.}

\begin{figure*}[t]
    \centering
    \begin{subfigure}[t]{0.49\textwidth}
        \centering
        \includegraphics[width=\textwidth]{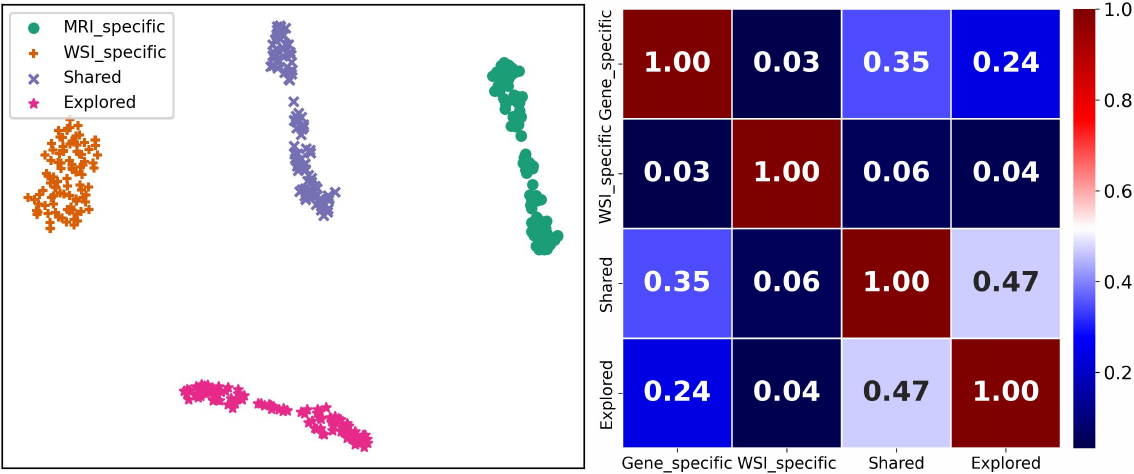}
        \caption{t-SNE and heatmap on LC dataset}
    \end{subfigure}
    \hfill
    \begin{subfigure}[t]{0.49\textwidth}
        \centering
        \includegraphics[width=\textwidth]{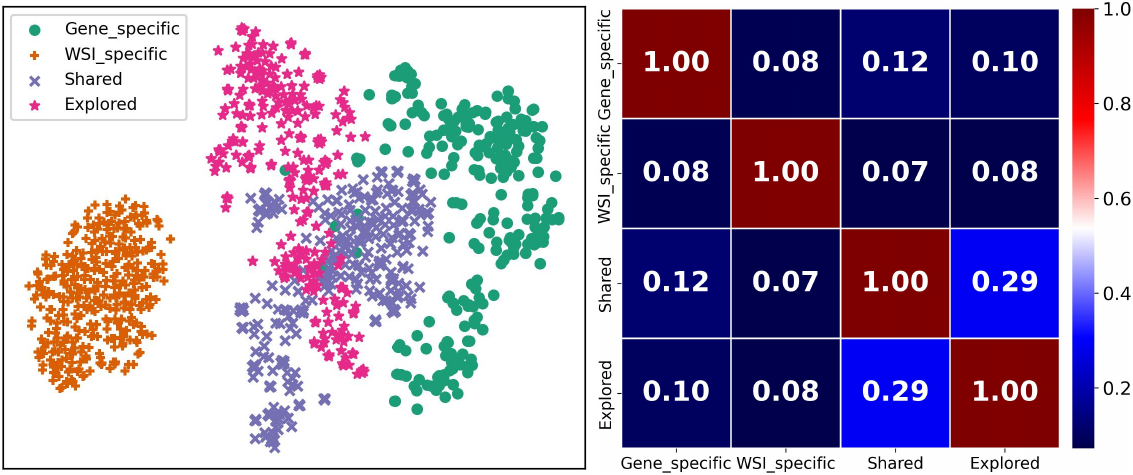}
        \caption{t-SNE and heatmap on BLCA dataset}
    \end{subfigure}
     \begin{subfigure}[t]{0.49\textwidth}
        \centering
        \includegraphics[width=\textwidth]{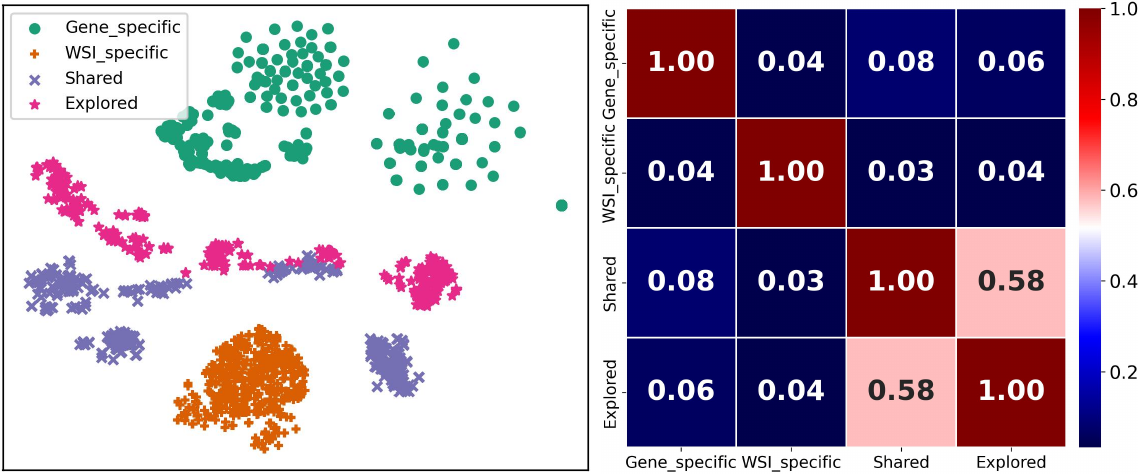}
        \caption{t-SNE and heatmap on UCEC dataset}
    \end{subfigure}
    \hfill
   \begin{subfigure}[t]{0.49\textwidth}
        \centering
        \includegraphics[width=\textwidth]{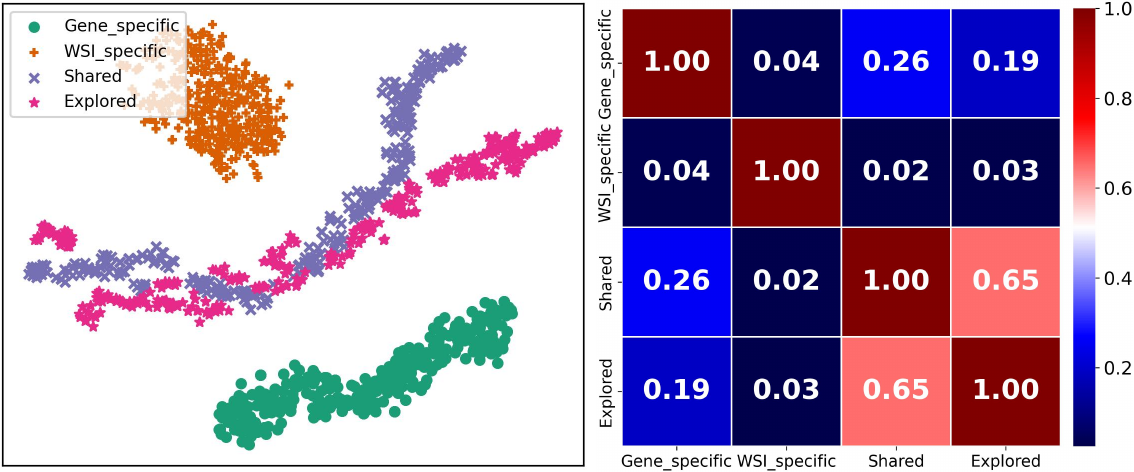}
        \caption{t-SNE and heatmap on LUAD dataset}
    \end{subfigure}
    \caption{Visualization results of decoupled features. In each sub-figure, the left part displays t-SNE visualization of the decoupled features, and the right part shows the heatmap, which displays the \rev{centered kernel alignment (CKA)} similarity of each pair-wise decoupled
    features.}
    \label{fig:vis_tsne}
\end{figure*}


\begin{figure*}[t]
    \centering
    \begin{subfigure}{0.24\textwidth}
        \centering
        \includegraphics[width=\textwidth]{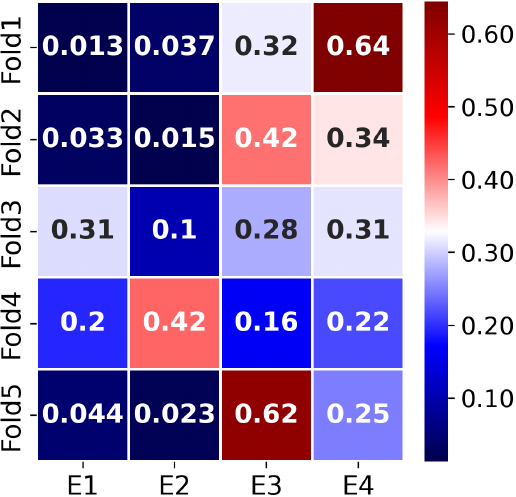}
        \caption{LC dataset}
    \end{subfigure}
    \hfill
    \begin{subfigure}{0.24\textwidth}
        \centering
        \includegraphics[width=\textwidth]{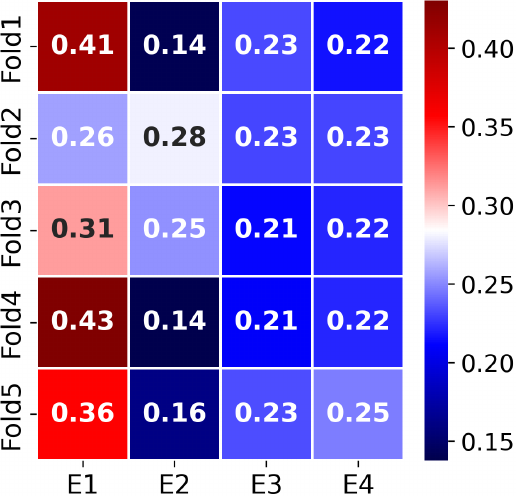}
        \caption{BLCA dataset}
    \end{subfigure}
    \hfill
    \begin{subfigure}{0.24\linewidth}
		\centering
		\includegraphics[width=\textwidth]{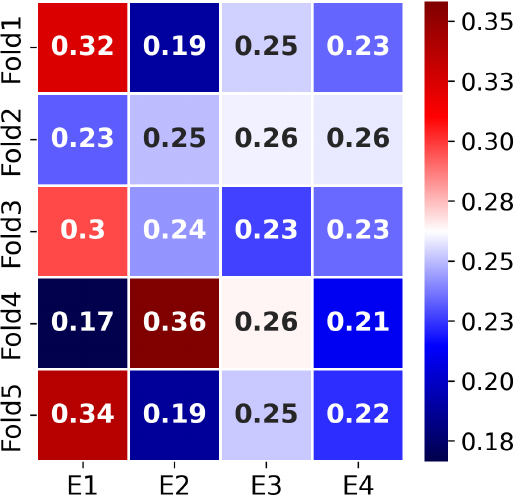}
		\caption{UCEC dataset}
	\end{subfigure}
    \hfill
	\begin{subfigure}{0.24\linewidth}
		\centering
		\includegraphics[width=\textwidth]{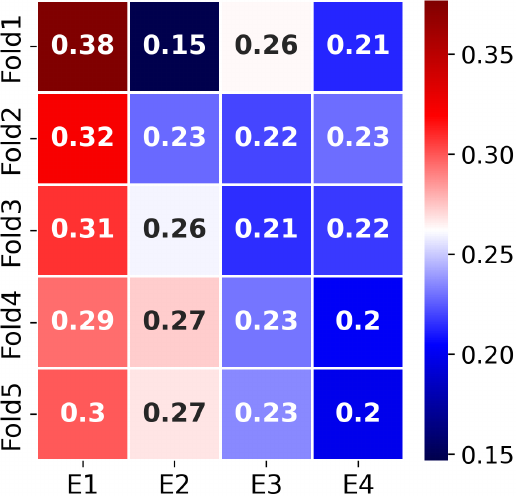}
		\caption{LUAD dataset}
    \end{subfigure}
      \caption{Visualization results of dynamic average weights from the output of MoE Gating unit. 
      }
    \label{fig:weight}
        \vspace{-2ex}
\end{figure*}


\begin{figure*}[t]
    \centering
    \begin{subfigure}{0.24\textwidth}
        \centering
        \includegraphics[width=0.98\textwidth]{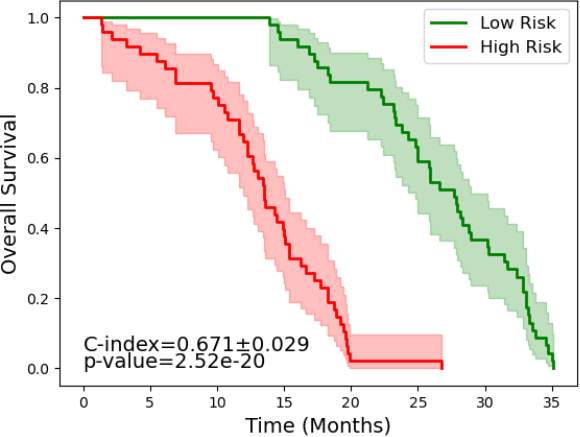}
        \caption{LC dataset}
    \end{subfigure}
    \hfill
    \begin{subfigure}{0.24\textwidth}
        \centering
        \includegraphics[width=0.98\textwidth]{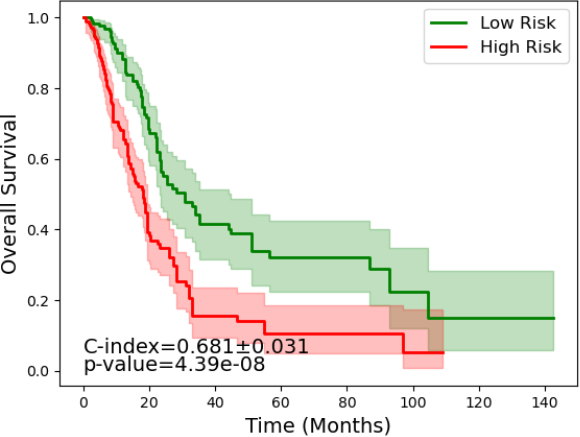}
        \caption{BLCA dataset}
    \end{subfigure}
    \hfill
    \begin{subfigure}{0.24\linewidth}
		\centering
		\includegraphics[width=0.98\linewidth]{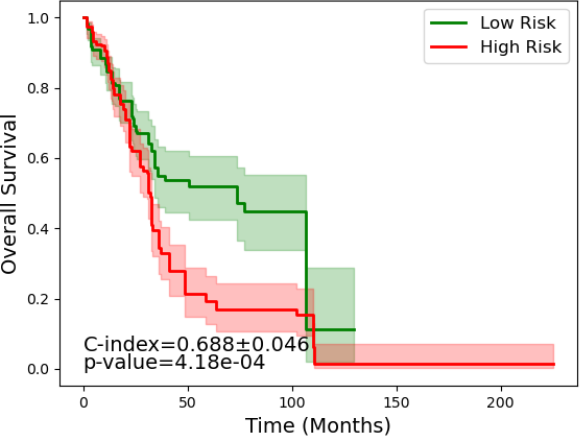}
		\caption{UCEC dataset}
	\end{subfigure}
    \hfill
	\begin{subfigure}{0.24\linewidth}
		\centering
		\includegraphics[width=1\linewidth]{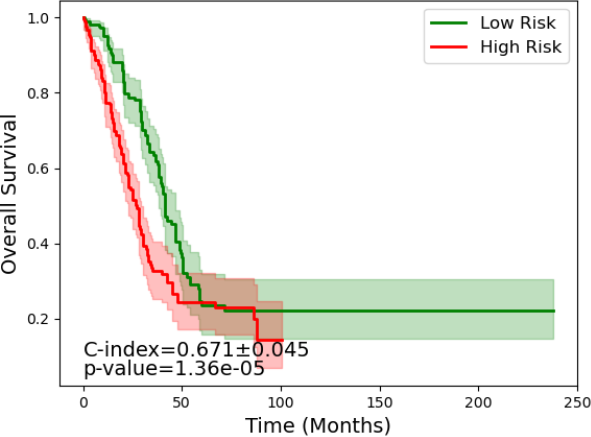}
		\caption{LUAD dataset}
	\end{subfigure}
       \caption{Visualization of Kaplan-Meier Analysis, where patient stratifications of low risk (green) and high risk (red) are presented. Shaded areas refer to the confidence intervals.
       }
    \label{fig:km}
        \vspace{-2ex}
\end{figure*}

\begin{figure*}[t]
    \centering
    \begin{subfigure}{0.24\textwidth}
        \centering
        \includegraphics[width=0.98\textwidth]{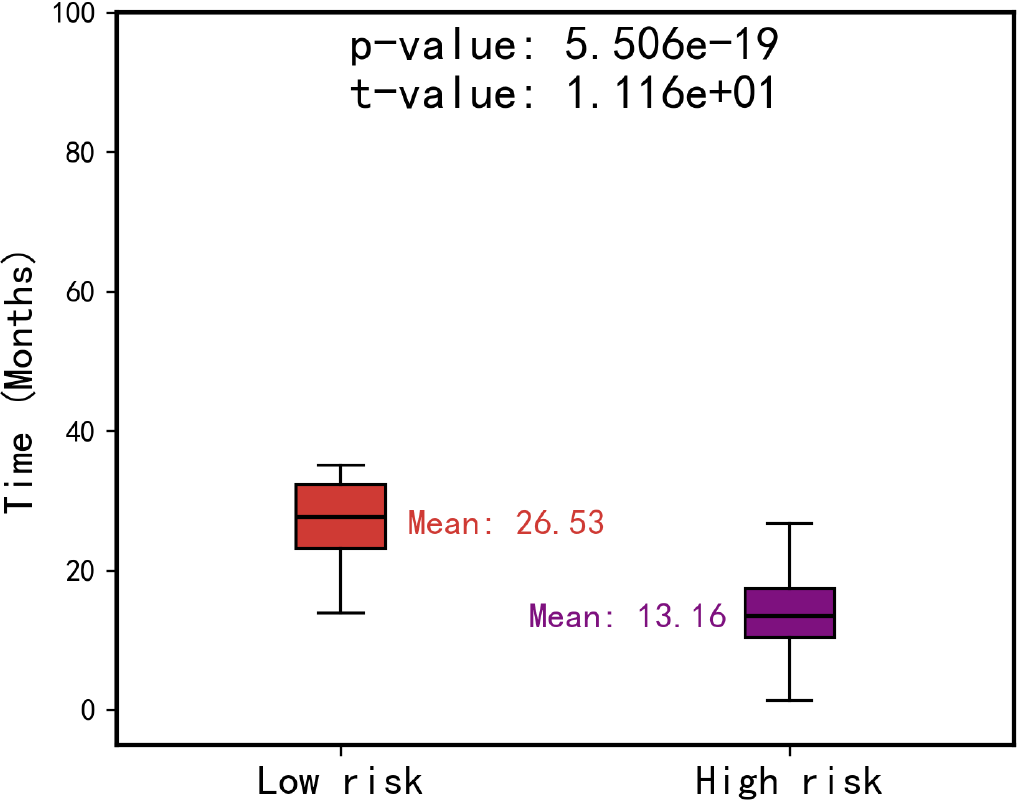}
        \caption{LC dataset}
    \end{subfigure}
    \hfill
    \begin{subfigure}{0.24\textwidth}
        \centering
        \includegraphics[width=0.98\textwidth]{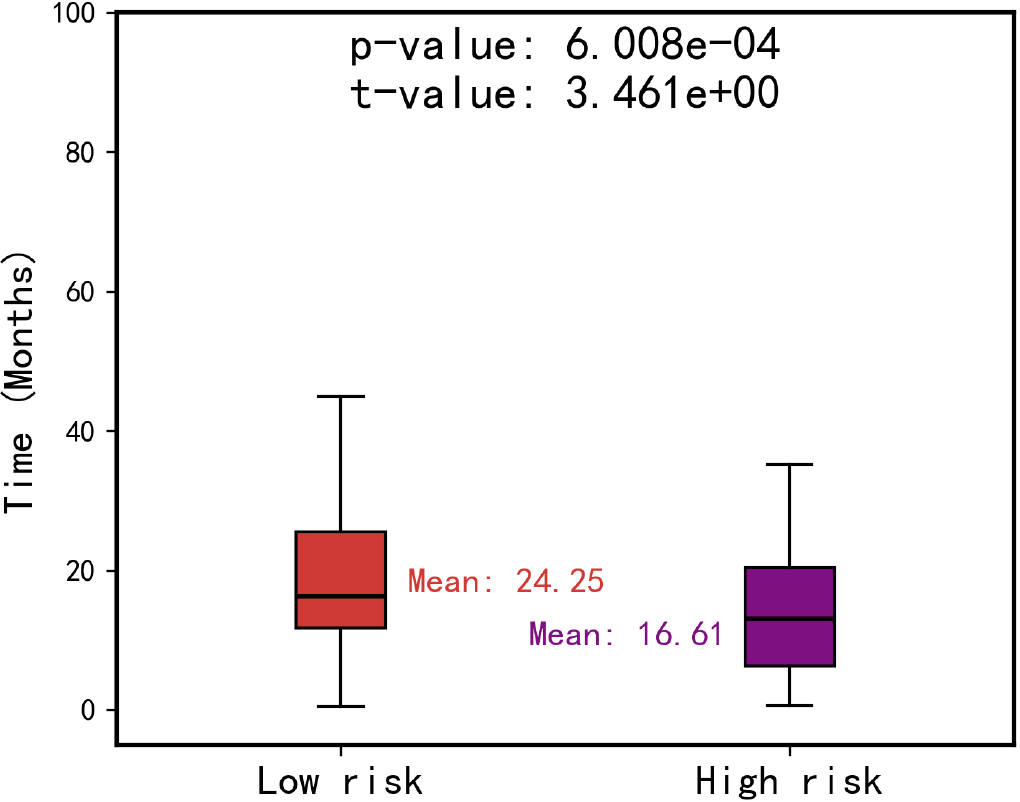}
        \caption{BLCA dataset}
    \end{subfigure}
    \hfill
    \begin{subfigure}{0.24\linewidth}
		\centering
		\includegraphics[width=0.98\linewidth]{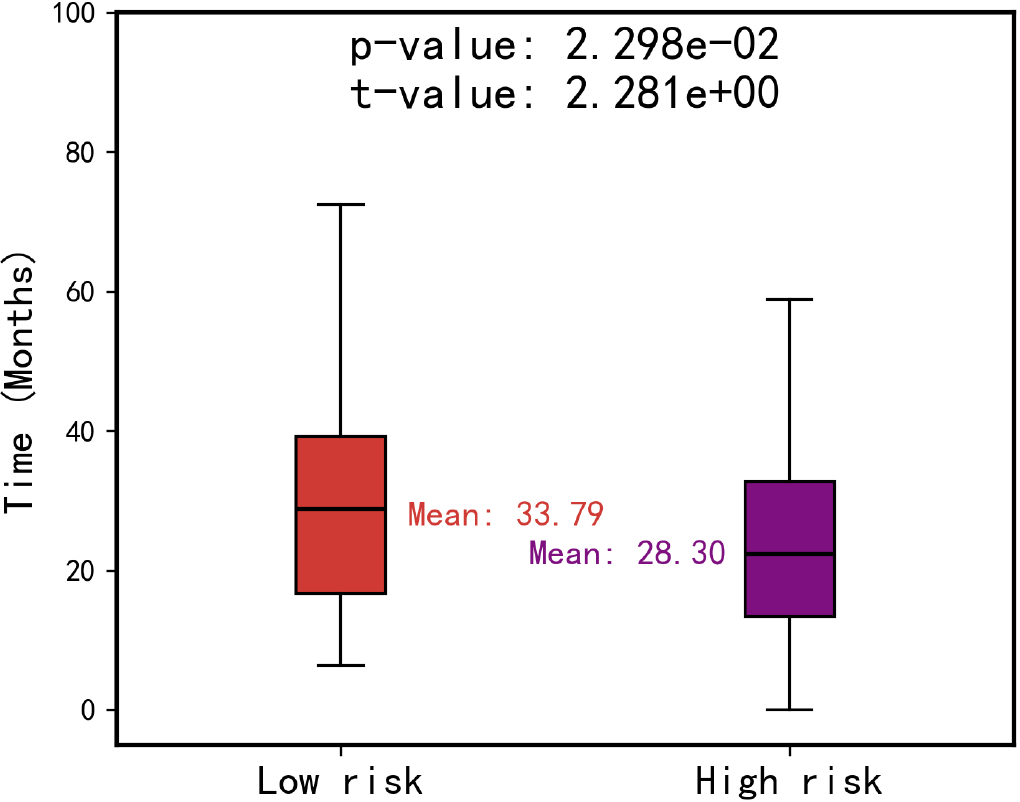}
		\caption{UCEC dataset}
	\end{subfigure}
    \hfill
	\begin{subfigure}{0.24\linewidth}
		\centering
		\includegraphics[width=1\linewidth]{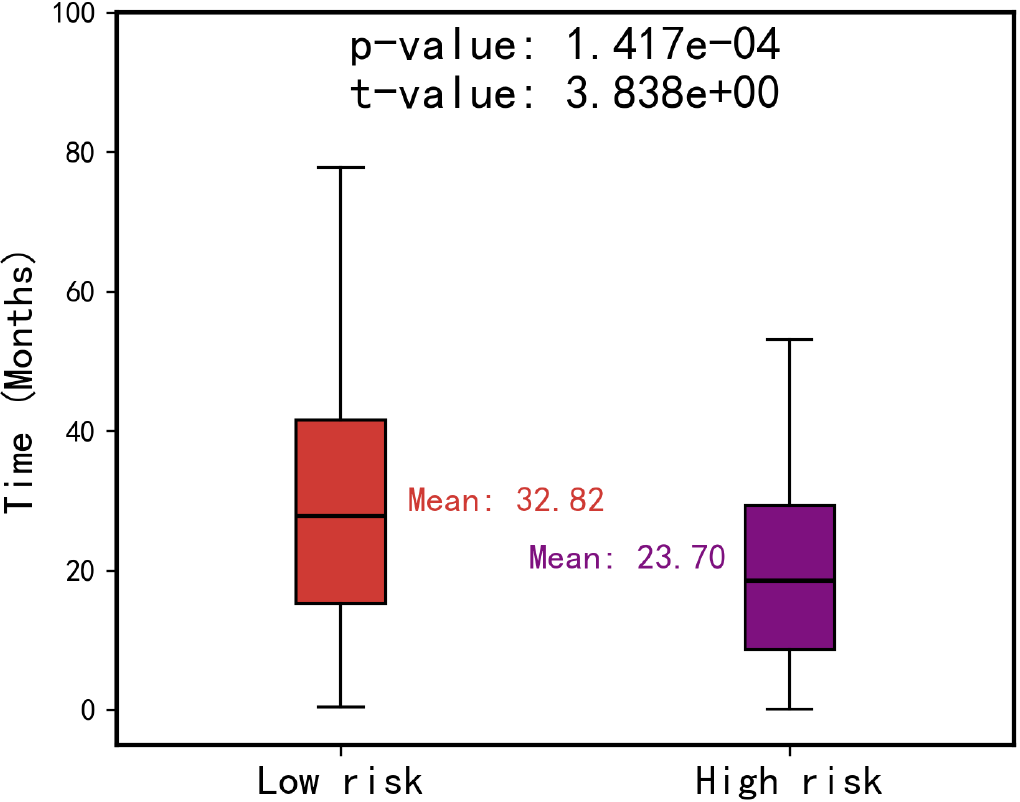}
		\caption{LUAD dataset}
	\end{subfigure}
       \caption{Visualization of T-test Analysis, where patient box-plots of low risk (red) and high risk (purple) are presented.
       }
    \label{fig:t_test}
        \vspace{-2ex}
\end{figure*}

\begin{figure*}[t]
    \centering
    \begin{subfigure}[t]{0.24\textwidth}
        \centering
        \includegraphics[width=\textwidth]{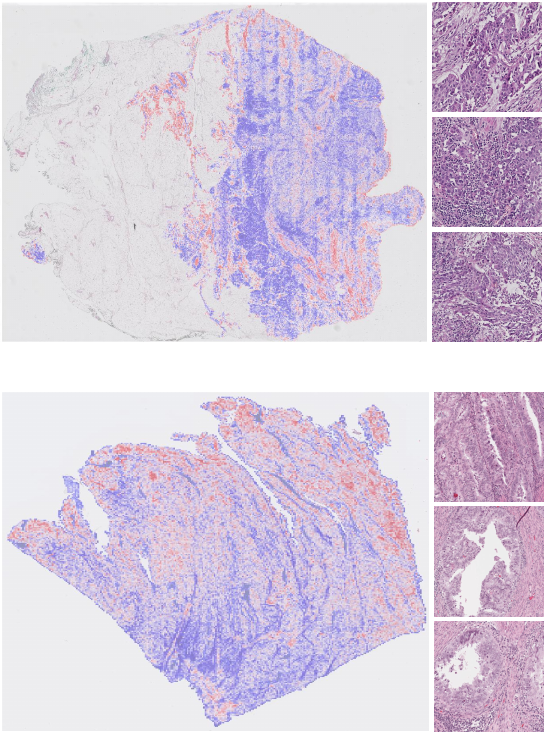}
        \caption{Modality-Shared Representation}
    \end{subfigure}
    \begin{subfigure}[t]{0.24\textwidth}
        \centering
        \includegraphics[width=\textwidth]{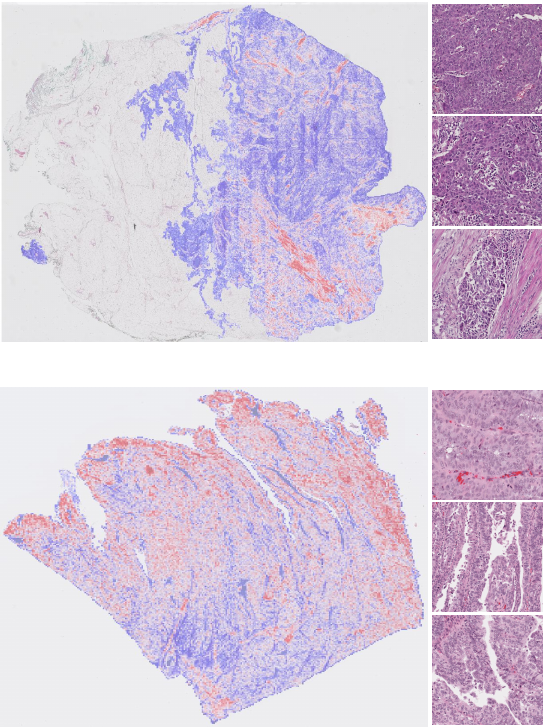}
        \caption{Modality-Explored Representation}
    \end{subfigure}
     \begin{subfigure}[t]{0.24\textwidth}
        \centering
        \includegraphics[width=\textwidth]{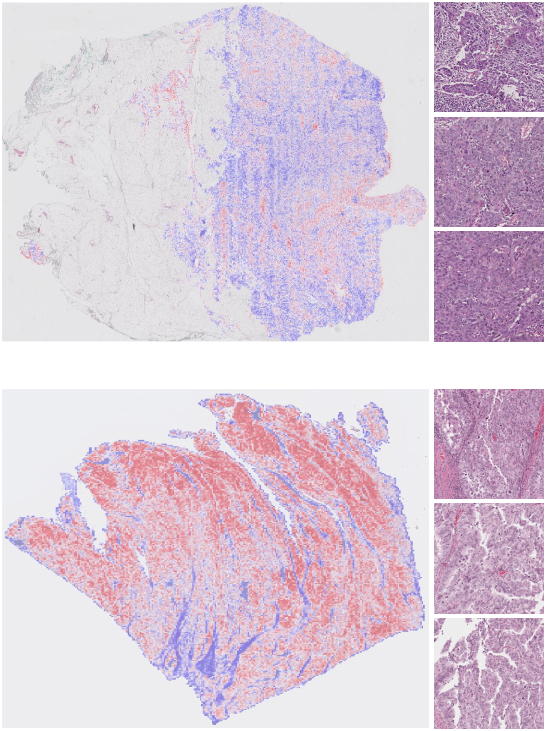}
        \caption{Pathology-Specific Representation}
    \end{subfigure}
   \begin{subfigure}[t]{0.24\textwidth}
        \centering
        \includegraphics[width=0.94\textwidth]{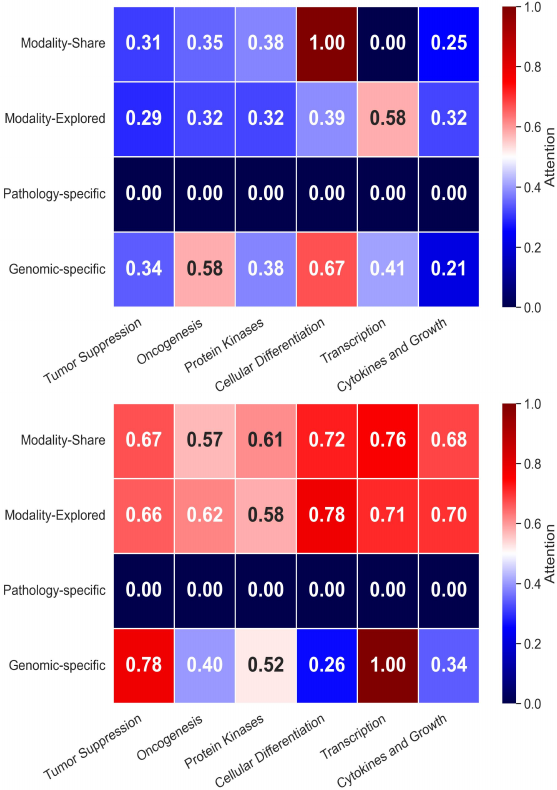}
        \caption{Attention on genomics}
    \end{subfigure}
    \caption{Attention visualization of different decoupled features on input modalities.}
    \label{fig:vis_wsi}
\end{figure*}

\subsection{Visualization Analysis}
Next, we visualize the distributions of decoupled features, weights in MoE, KM curves and attention analysis of decoupled features on LC, BLCA, LUAD, UCEC datasets. 

\noindent\textbf{Decoupled Features Visualization.} In Fig.~\ref{fig:vis_tsne}, we use \revm{t-distributed Stochastic Neighbor Embedding (t-SNE)}~\cite{hinton2002stochastic} to visualize the distribution of decoupled features. \rev{We also use Centered Kernel Alignment (CKA) method~\cite{kornblith2019similarity} to analyze the correlation between the decoupled features.} The green, orange, purple, and pink points represent the dimensionality-reduced results of modality-specific, modality-shared, and modality-explored features, respectively. These distributions differ from each other, with modality-shared and modality-explored features positioned between the modality-specific feature points. This indicates that modality-shared and modality-explored features possess intermediary properties across different modalities. \revm{Details of CKA calculation are provided in the Supplementary Materials}

\noindent\textbf{Weights Visualization in MoE Module.} We compute the average fusion weights output by the Gating unit in MoE module on four test datasets in Fig.~\ref{fig:weight}. In all cases, each expert network exhibits different output weights. This variation reflects the concept of dynamic fusion, in which each expert network focuses on distinct aspects with varying levels of importance. The weight dynamic fusion enables the model to leverage comprehensive feature information.

\noindent\textbf{Kaplan-Meier Analysis.} The Kaplan-Meier (KM) test \cite{kaplan1958nonparametric} is a nonparametric statistical method used to estimate survival probabilities and analyze time-to-event data. 
We use the median survival time to divide all patients into high- and low-risk groups, represented as red and green curves, respectively.
As shown in Fig.~\ref{fig:km}, the p-values achieved by our framework are significantly lower than 0.05 on all four datasets, demonstrating statistically significant discrimination between high- and low-risk groups.

\noindent\textbf{T-test Analysis.} \rev{The T-test is a statistical hypothesis test used to determine whether there is a significant difference between the means of two groups. 
In the t-test analysis, the t-value quantifies the standardized degree of difference between the group means, while the p-value measures the probability of the null hypothesis being ture (\ie, the likelihood that our results occurring by chance). 
We use the median survival time to divide all patients into high- and low-risk groups for t-test analysis. As shown in Fig.~\ref{fig:t_test}, on four datasets, our framework effectively distinguishes between the two groups with favorable p-values and t-values (all p-values $<$ 0.05). 
}

\noindent\textbf{Attention Analysis of Decoupled Features.} To further illustrate the differences among various decoupled features, we select two examples from TCGA dataset and visualize the attention scores on input data in Fig.~\ref{fig:vis_wsi}. 
The sub-figures (a)-(c) demonstrate the attention regions of modality-shared, modality-explored, and modality-specific features on input pathology images, while the last sub-figure (d) presents the attention scores of decoupled features across six gene pathways.
In these samples, for pathology images, there are less or more differences in the regions focused by different decoupled features, proving that these decoupled features can prompt the model to capture more comprehensive information. Additionally, we select several high-scoring local patches for magnified visualization, and these regions are validated by a pathologist. The identified patches exhibit characteristic tumor features such as cellular pleomorphism, enlarged nuclei, increased nuclear-to-cytoplasmic ratio, and pathological mitotic figures, demonstrating that the model can accurately recognize regions with discriminative pathological information. 
For genomic profiles, it can be found that different decoupled features have different importance for six genomic subsequences.  
We also observe that the attention weights between the pathology-specific feature and the six gene pathways are zero. This is expected, as the pathology-specific feature is designed to encode information unique to the WSI modality, such as architectural atypia and nuclear pleomorphism. Our proposed decoupling module successfully separates the pathology-specific feature from other features, as further supported by Fig.~\ref{fig:vis_tsne}. Therefore, the pathology-specific feature exhibits strong non-correlation with the six gene pathway-related features.
The above results imply that two types of multimodal interactions are necessary for effective survival analysis. \revm{For more visualization examples and the calculation method of attention score, please refer to the Supplementary Materials.}

\section{Conclusions}
In this paper, we proposed DeReF for multimodal cancer survival prediction. Two strategies were devised within this framework, including regional cross-attention and random feature reorganization. Specifically, first, before MoE fusion, we randomly reorganized decoupled features, which is capable of reducing the over-reliance of expert networks on fixed feature recombinations, enhancing the generalization of the expert networks. It could also address the problem of information closure and capture more interaction information among decoupled features. Second, we proposed a regional cross-attention network for extracting modality-shared and modality-explored decoupled features. It is able to fully analyze inter- and intra-relationships of modality features to extract better decoupled representations. Results on four datasets showed that DeReF performed well in cancer survival analysis tasks.

\why{Despite the proven effectiveness of our work, there are several improvements for the future. First, while the feature reorganization algorithm reduces the model's reliance on fixed feature combinations to some extent, the current algorithm operates based on predefined segment lengths. In future work, we will study completely random shuffle algorithm, where random segments are reorganized at random positions. We hypothesize that this approach could completely eliminate the model's dependence on fixed feature combinations, thereby achieving better performance. 
}



\bibliographystyle{IEEEtran}  
\bibliography{main.bib}

\end{document}